\documentclass[twoside,11pt]{article}
\usepackage{jmlr2e}
\usepackage{lmodern}
\usepackage[T1]{fontenc}
\usepackage[utf8]{inputenc}
\usepackage{microtype}
\usepackage{csquotes}
\usepackage{pdfpages}
\usepackage{float}
\usepackage{grffile}
\usepackage{amsmath}
\usepackage{subdepth}
\usepackage{accents}
\newcommand\thickbar[1]{\accentset{\rule{.4em}{.8pt}}{#1}}

\newcommand\thickubar[1]{\underaccent{\rule{.45em}{.8pt}}{#1}}

\newcommand\indep{\protect\mathpalette{\protect\independenT}{\perp}}
\def\independenT#1#2{\mathrel{\rlap{$#1#2$}\mkern2mu{#1#2}}}
\newcommand{\+}[1]{\ensuremath{\mathbf{#1}}}

\newcommand{\lineref}[2]{A\ref{#1}:\ref{#2}}
\def\independenT#1#2{\mathrel{\rlap{$#1#2$}\mkern2mu{#1#2}}}

\usepackage{algorithm} 
\usepackage{algorithmicx} 
\usepackage[noend]{algpseudocode}

\algnewcommand\algorithmicinput{\textbf{INPUT:}}
\algnewcommand\INPUT{\item[\algorithmicinput]}
\algnewcommand\algorithmicoutput{\textbf{OUTPUT:}}
\algnewcommand\OUTPUT{\item[\algorithmicoutput]}
\usepackage{listings}

\jmlrheading{18}{2017}{1-30}{4/16; Revised 3/17}{4/17}{16-166}{Santtu Tikka and Juha Karvanen}

\ShortHeadings{Simplifying Probabilistic Expressions}{Tikka and Karvanen}
\firstpageno{1}

\begin{document}

\title{Simplifying Probabilistic Expressions in Causal Inference}

\author{\name Santtu Tikka \email santtu.tikka@jyu.fi \\
              \name Juha Karvanen \email juha.t.karvanen@jyu.fi \\
              \addr Department of Mathematics and Statistics \\
              P.O.Box 35 (MaD) FI-40014 University of Jyvaskyla, Finland}

\editor{Peter Spirtes}

\maketitle

\begin{abstract}%
Obtaining a non-parametric expression for an interventional distribution is one of the most fundamental tasks in causal inference. Such an expression can be obtained for an identifiable causal effect by an algorithm or by manual application of do-calculus.
Often we are left with a complicated expression which can lead to biased or inefficient estimates when missing data or measurement errors are involved.

We present an automatic simplification algorithm that seeks to eliminate symbolically unnecessary variables from these expressions by taking advantage of the structure of the underlying graphical model. Our method is applicable to all causal effect formulas and is readily available in the R package causaleffect.
\end{abstract}


\section{Introduction} \label{sect:intro} 
Symbolic derivations resulting in complicated expressions are often encountered in many fields working with mathematical notation. These expressions can be derived manually or they can be outputs from a computer algorithm. In both cases, the expressions may be correct but unnecessarily complex in a sense that some unrecognized identities or properties would lead to simpler expressions.

We will consider simplification in the context of causal inference in graphical models \citep{pearl09}. Advances in causal inference have led to algorithmic solutions to problems such as identifiability of causal effects and conditional causal effects \citep{huang06, shpitser06, shpitser06cond}, $z$-identifiability \citep{Bareinboim:zidentifiability}, transportability and meta-transportability \citep{bareinboim2013general, Bareinboim:metatransportability} among others. The aforementioned algorithmic solutions operate symbolically on the joint distribution of the variables of interest and return expressions for the desired queries. These algorithms have been previously implemented in the R package causaleffect \citep{tikka17}. Another implementation of an identifiability algorithm can be found in the CIBN software by Jin Tian and Lexin Liu freely available from \url{http://web.cs.iastate.edu/~jtian/Software/CIBN.htm}. However, the algorithms themselves are imperfect in a sense that they often output an expression that is complicated and far from ideal. The question is whether there exists a simpler expression that is still a solution to the original problem. 

Simplification of expressions may provide significant benefits. First, a simpler expression can be understood and reported more easily. Second, evaluating a simpler expression will be less of a computational burden due to reduced dimensionality of the problem. Third, in situations where estimation of causal effects is of interest and missing data is a concern, eliminating variables with missing data from the expression has clear advantages. The same applies to variables with measurement error.

We begin with presenting in Section~\ref{sect:expr} a general form of probabilistic expressions that are often encountered in causal inference. In this paper probabilistic expressions are formed by products of non-parametric conditional distributions of some variables and summations over the possible values of these variables. Simplification in this case is the process of eliminating terms from these expressions by carrying out summations. As our expressions correspond to causal effects, the expressions themselves take a specific form.

Causal models are typically associated with a directed acyclic graph (DAG) which represents the functional relationships between the variables of interest. In situations where the joint distribution is faithful, meaning that no additional conditional independences are generated by the joint distribution \citep{spirtes00}, the conditional independence properties of the variables can be read from the graph itself through a concept known as d-separation \citep{geiger90}. We will use d-separation as our primary tool for operating on the probabilistic expressions. The reader is assumed to be familiar with a number of graph theoretic concepts that are explained for example in \citep{koller09} and used throughout the paper.

Our simplification procedure is built on the definition of simplification sets, which is presented in Section~\ref{sect:simp}. We continue by introducing a sound and complete simplification algorithm for probabilistic expressions defined in Section~\ref{sect:expr} for which these simplification sets exist. The algorithm takes as an input the expression to be simplified and the graph induced by the underlying causal model, and proceeds to construct a joint distribution of the variables contained in the expression by using the d-separation criteria. Higher level algorithms that use this simplification procedure are presented in Section~\ref{sect:algo}. These include an algorithm for the simplification of a nested expression and an algorithm for the simplification of a quotient of two expressions. Section~\ref{sect:exam} contains examples on the application of these algorithms. We have also updated the causaleffect R-package to automatically apply these simplification procedures to causal effect expressions.

As a motivating example we present an expression of a causal effect given by the ID algorithm of \citet{shpitser06} that can be simplified. The complete derivation of this effect can be found in Appendix~\ref{app:derivation}.
\begin{figure}[h]
  \centering
  \includegraphics[width=0.40\textwidth]{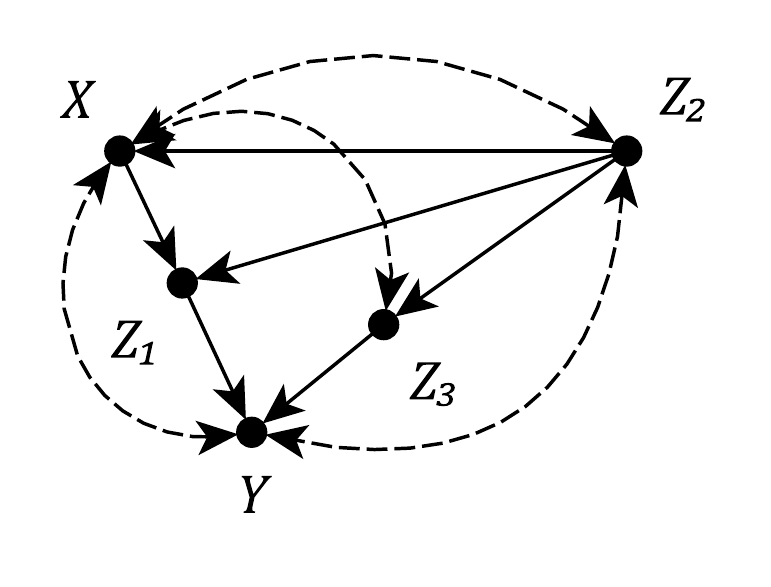}
  \caption{A graph for the introductory example on simplification.}
  \label{fig:grfG}
\end{figure}
\noindent
The causal effect of $X$ on $Z_1, Z_2, Z_3$ and $Y$ is identifiable in the graph of Figure~\ref{fig:grfG} and application of the ID algorithm gives
\begin{align*}
 P(Z_1| Z_2,X)P(Z_3| Z_2)\frac{\sum_{X}P(Y| Z_2,X,Z_3,Z_1)P(Z_3| Z_2,X)P(X| Z_2)P(Z_2)}{\sum_{X,Y}P(Y| Z_2,X,Z_3,Z_1)P(Z_3| Z_2,X)P(X| Z_2)P(Z_2)} \times \\ \sum_{X,Z_3,Y} P(Y| Z_2,X,Z_3,Z_1)P(Z_3| Z_2,X)P(X| Z_2)P(Z_2).
\end{align*}
It turns out that there exists a significantly simpler expression,
\begin{equation} \label{eq:motivation}
P(Z_1|Z_2,X)P(Z_2) \sum_{X} P(Y|Z_2,X,Z_3,Z_1)P(Z_3|Z_2,X)P(X|Z_2),
\end{equation}
for the same causal effect. This expression can be obtained without any knowledge of the underlying model by using standard probability manipulations. However, this requires that a favorable choice is made for the ordering of the nodes of the graph in the ID algorithm. In the case that we had chosen an ordering where $Z_1$ precedes $Z_3$, the term for $Z_3$ would instead be $P(Z_3|Z_2, Z_1, X)$ and simplification would require knowledge about the underlying graph. We will take another look at this example later in Section~\ref{sect:exam} where we describe in detail how our procedure can be used to find expression (\ref{eq:motivation}). 

Our simplification procedure is different from the well-known exact inference method of minimizing the amount of numerical computations when evaluating expressions for conditional and marginal distributions by changing the order of summations and multiplications in the expression. Variants of this method are known by different names depending on the context, such as Bayesian variable elimination \citep{koller09} and the sum-product algorithm \citep{bishop06} which is a generalization of belief propagation \citep{Pearl:1988, lauritzen98}. Efficient computational methods exist for causal effects as well, such as \citep{shpitser11}. The general principle is the same in all of the variants, and no symbolic simplification is performed. 

In our setting simplification can be defined explicitly but in general it is difficult to say what makes one expression simpler than another. \citet{carette04} provides a formal definition for simplification in the context of Computer Algebra Systems (CAS) that operate on algebraic expressions. Modern CAS systems such as Mathematica \citep{mathematica} and Maxima \citep{maxima} implement techniques for symbolic simplification. \citet{Bailey2014120} and references therein discuss simplification techniques in CAS systems further. However to the best of our knowledge, the symbolic simplification procedures for probabilistic expressions described in this paper have neither been given previous attention nor implemented in any existing system.

\section{Probabilistic Expressions} \label{sect:expr}
Every expression that we consider is defined in terms of a set of variables $\+ W$. As we are interested in probabilistic expressions, we also assume a joint probability distribution $P$ for the variables of $\+ W$. The most basic of expressions are called atomic expressions which will be the main focus of this paper.

\begin{definition}[Atomic expression] \label{def:atomic} Let $\+ W$ be a set of $p$ discrete random variables and let $P$ be any joint distribution of $\+ W$. An atomic expression is a pair
\[ A = A[\+W] = \langle \+ T, \+ S \rangle, \]
where
\begin{enumerate}
\item $\+ T$ is a set of pairs $\{ \langle V_1, \+ C_1 \rangle, \ldots, \langle V_n, \+ C_n \rangle \}$ such that for each $V_i$ and $\+ C_i$ it holds that $V_i \in \+ W$, $\+ C_i \subseteq \+ W$, $V_i \not\in \+ C_i$ and $V_i \neq V_j$ for $i \neq j$.
\item $\+ S$ is a set $\{S_1, \ldots ,S_m\} \subseteq \+ W$ such that for each $i = 1,\ldots,m$ it holds that $S_i = V_j$ for some $j \in \{1,\ldots,n\}$.
\end{enumerate}
The value of an atomic expression $A$ is 
\[ P_A =  \sum_{\+ S} \prod_{i=1}^n P(V_i|\+ C_i). \]
\end{definition}
The probabilities $P(V_i|\+ C_i)$ are referred to as the terms of the atomic expression. A term $P(V_i|\+ C_i)$ is said to contain a variable $V$ if $V_i = V$ or $V \in \+ C_i$. A term for a variable $V$ refers to a term $P(V|\cdot)$. We also use the shorthand notation $V[A] := \{V_1,\ldots V_n\}$. As $\+ S$ is a set, we will only sum over a certain variable once.
All variables are assumed to be univariate and discrete for clarity, but we may also consider multivariates and situations where some of the variables are continuous and the respective sums are interpreted as integrals instead. 

As an example we will construct an atomic expression describing the following formula 
\[ \sum_{X}P(Y| Z_2,X,Z_3,Z_1)P(Z_3| Z_2,X)P(X| Z_2)P(Z_2), \]
which is a part of the motivating example in the introduction. We let $\+ W = \{X, Y, Z_1, Z_2, Z_3\}$, which is the set of nodes of the graph of Figure~\ref{fig:grfG}. The sets $\+ T$ and $\+ S$ can now be defined as 
\[ \{ \langle Y, \{Z_2,X,Z_3,Z_1\} \rangle, \langle Z_3 , \{Z_2, X\} \rangle, \langle X , \{Z_2\} \rangle , \langle  Z_2, \emptyset \rangle\} \quad \textrm{ and } \quad \{X\},\] respectively. Next we define a more general probabilistic expression.

\begin{definition}[Expression] \label{def:expr} Let $\+ W$ be a set of $p$ variables and let $P$ be the joint distribution of $\+ W$. An expression is a triple
\[ B = B[\+W,n,m] = \langle \+ B, \+A, \+ S \rangle, \]
where
\begin{enumerate}
\item $\+ S$ is a subset of $\+ W$.
\item For $m > 0$, $\+ A$ is a set of atomic expressions 
\[ \{\langle \+ T_1, \+ S_1 \rangle, \ldots, \langle \+ T_m, \+ S_m \rangle\}. \] 
 If $m = 0$ then $\+ A = \emptyset$. 
\item For $n > 0$, $\+ B$ is a set of expressions
\[ \{B_1[\+ W_1,n_1,m_1], \ldots ,B_n[\+ W_n,n_n,m_n]\} \] such that $\+ W_i \subseteq \+ W$, $n_i < n, m_i < m$ for all $i = 1,\ldots,n$. If $n = 0$ then $\+ B = \emptyset$. 
\end{enumerate}
The value of an expression $B$ is
\[  P_B =  \sum_{\+ S} \prod_{i=1}^n P_{B_i} \prod_{j=1}^m P_{A_j}, \]
where an empty product should be understood as being equal to 1.
\end{definition}

The recursive definition ensures the finiteness of the resulting expression by requiring that each sub-expression has fewer sub-expressions of their own than the expression above it. A single value might be shared by multiple expressions, as the terms of the product in the value of the expression are exchangeable. Expressions $B_1[\+W,n_1,m_1]$ and $B_2[\+W,n_2,m_2]$ are equivalent if their values $P_{B_1}$ and $P_{B_2}$ are equal for all $P$. Equivalence is defined similarly for atomic expressions. Every expression is formed by nested atomic expressions by definition. Because of this, we focus on the simplification of atomic expressions.

As an example we construct an expression for the causal effect formula (\ref{eq:motivation}). We define $\+ W := \{X, Y, Z_1, Z_2, Z_3\}$ and let the sets $\+ B$ and $\+ S$ be empty. We define the set $\+ A$ to consist of three atomic expressions $A_1, A_2$ and $A_3$ defined as follows
\begin{align*}
  A_1 &= \langle \{ \langle Z_1, \{ Z_2, X \} \rangle \}, \emptyset \rangle, \\
  A_2 &= \langle \{ \langle Z_2, \emptyset \} \rangle \}, \emptyset \rangle, \\
  A_2 &= \langle \{ \langle Y, \{Z_2,X,Z_3,Z_1\} \rangle, \langle Z_3 , \{Z_2, X\} \rangle, \langle X , \{Z_2\} \rangle , \langle  Z_2, \emptyset \rangle\}, \{X\} \rangle.
\end{align*}

In the context of probabilistic graphical models, we are provided additional information about the joint distribution of the variables of interest in the form of a DAG. As we are concerned on the simplification of the results of causal effect derivations in such models, the general form of the atomic expressions can be further narrowed down by using the structure of the graph and the ordering of vertices called a topological ordering.

\begin{definition}[Topological ordering]
Topological ordering $\pi$ of a DAG $G = \langle \+ W, \+ E \rangle$ is an ordering of its vertices, such that if $X$ is an ancestor of $Y$ in $G$ then $X < Y$ in $\pi$. 
\end{definition}

The symbol $V_j^\pi$ is used to denote the subset of vertices of $G$ that are less than $V_j$ in $\pi$. For sets we may define $\+ V^\pi$ to contain those vertices of $G$ that are less than every vertex of $\+ V$ in $\pi$. Consider a DAG $G = \langle \+ W, \+ E \rangle$ and a topological ordering $\pi$ of its vertices. We use the notation $\pi(\cdot)$ to denote indexing over the vertex set $\+ W$ of $G$ in the ordering given by $\pi$, that is $V_{\pi(1)} >  V_{\pi(2)} > \cdots > V_{\pi(m)}$ where $m = |\+ W|$. For any atomic expression $A[\+ V] = \langle \+ T, \+ S \rangle$ such that $\+ V \subseteq \+ W$ we also define the induced ordering $\omega$. This ordering is an ordering of the variables in $\+ V$ such that if $X > Y$ in $\omega$ then $X  > Y$ also in $\pi$. From now on in this paper, any indexing over the variables of an atomic expression will refer to the induced ordering of the set $\+ V$ when $\pi$ is given, i.e $V_1 > V_2 > \cdots > V_n$ in $\omega$. In other words, $\omega$ is obtained from $\pi$ by leaving out variables that are not contained in $A$.

The ID algorithm performs the so-called C-component factorization. These components are subgraphs of the original graph where every node is connected by a path consisting entirely of bidirected edges. The resulting expressions of these factors serve as the basis for our simplification procedure. 

\begin{definition}[Topological consistency] Let $G^\prime$ be a DAG with a subgraph $G = \langle \+ W, \+ E \rangle$ and let $\pi$ be a topological ordering of the vertices of $G$. An atomic expression $A[\+ W] = \langle \+ T, \+ S \rangle$ is topologically consistent (or $\pi$-consistent for short) if \[  An(V_i)_{G} \subseteq \+ C_i \subseteq V_i^\pi \textrm{ for all } i = 1,\ldots,n. \]
\end{definition} 
Here $An(V_i)_G$ denotes the ancestors of $V_i$ in $G$. To motivate this definition we note that the outputs of the algorithms of \citet{shpitser06,shpitser06cond} can always be represented by using products and quotients of topologically consistent atomic expressions. An expression is topologically consistent when every atomic expression contained by it is topologically consistent with respect to a topological ordering of a subgraph. We provide a proof for this statement in Appendix~\ref{app:consistent}. This also shows that any manual derivation of a causal effect can always be represented by a topologically consistent expression. The assumption that $ An(V_i)_{G} \subseteq \+ C_i$ is not necessary for the simplification to be successful. This assumption is used to speed up the performance of our procedure in Section~\ref{sect:simp}.

\section{Simplification} \label{sect:simp}
Simplification in our context is the procedure of eliminating variables from the set of variables that are to be summed over in expressions. In atomic expressions, a successful simplification in terms of a single variable should result in another expression that holds the same value, but with the respective term eliminated and the variable removed from the summation. As we are interested in causal effects, we consider only simplification of topologically consistent atomic expressions. 

Our approach to simplification is that the atomic expression has to represent a joint distribution of the variables present in the expression to make the procedure feasible. The question is whether the expression can be modified to represent a joint distribution. Before we can consider simplification, we have to define this property explicitly.


\begin{definition}[Simplification sets] \label{def:simpsets} Let $G^\prime$ be a DAG and let $G$ be a subgraph  of $G^\prime$ over a vertex set $\+ W$ with a topological ordering $\pi$. Let $A[\+ W] = \langle \+ T, \+ S \rangle$, where $\+ T = \{\langle V_1,\+ C_1, \rangle, \ldots, \langle V_n, \+ C_n \rangle \}$, be a $\pi$-consistent atomic expression and let $V_j \in \+ S$. Suppose that $V_{\pi(p)} = V_j$ and that $V_{\pi(q)} = V_1$ and let $\+ M$ be the set
\[  \{ U \in \+ W \mid U \not\in V[A], V_{\pi(q)} > U > V_{\pi(p)} \}. \]
If there exists a set $\+ D \subset V_j^\pi$ and the sets $\+ E_U \subseteq \+ W$ for all $U \in \+ M$ such that the conditional distribution of the variables $V_{\pi(p)}, \ldots, V_{\pi(q)}$ can be factorized as
\begin{align} \label{eq:factorization}
P(V_{\pi(p)},\ldots, V_{\pi(q)}|\+ D) = \prod_{U \in \+ M} P(U | \+ E_U) \prod_{V_i \geq V_j} P(V_i | \+ C_i),  
\end{align}
and
\begin{equation} \label{eq:independence}
(U \indep V_j | \+ E_U \setminus \{V_j\})_{G^\prime} \textrm{ for all } U \in \+ M.
\end{equation}
then the sets $\+ D$ and $\+ E_U, U \in \+ M$ are the simplification sets of $A$ with respect to $V_j$.
\end{definition}

This definition is tailored for the next result that can be used to determine the existence of a simpler expression when simplification sets exist. Afterwards we will show how this result can be applied in practice via an example. The definition characterizes $\pi$-consistent atomic expressions that represent joint distributions. It is apparent that simplifications sets are not always unique, which can lead to different but still simpler expressions. Henceforth the next result considers simplification in terms of a single variable. The proof is available in Appendix~\ref{app:simpl}.

\begin{theorem}[Simplification] \label{thm:simpl} Let $G^\prime$ be a DAG and let $G$ be a subgraph  of $G^\prime$ over a vertex set $\+ W$ with a topological ordering $\pi$. Let $A[\+ W] = \langle \+ T, \+ S \rangle$ be a $\pi$-consistent atomic expression and let $\+ D$ and $\+ E_U, U \in \+ M$ be its simplification sets with respect to a variable $V_j \in \+ S$. Then there exist an expression $A^\prime[\+ W \setminus \{V_j\}] = \langle \+ T^\prime, \+ S^\prime \rangle$ such that
$V_j \not\in \+S^\prime$, $P_{A} = P_{A'}$ and no term in $A^\prime$ contains $V_j$.
\end{theorem}

Note that even if $\+ M = \emptyset$ in Definition~\ref{def:simpsets}, the existence of simplification sets still requires that $\prod_{V_i \geq V_j} P(V_i | \+ C_i)$ = $P(V_{j},\ldots, V_{1}|\+ D)$. In many cases there exists variables $U \in \+ M$ such that the expression does not contain a term for $U$. Condition \eqref{eq:factorization} of Definition~\ref{def:simpsets} guarantees that if these terms were contained in the expression it would represent a joint distribution. Our goal is thus to introduce these terms into the original expression temporarily, carry out the desired summation, and finally remove the added terms. This can only be achieved if the variables in the set $\+ M$ are conditionally independent of the variable currently being summed over, hence the assumption $(U \indep V_j|\+ E_U \setminus \{V_j\})_{G^\prime}$ of condition \eqref{eq:independence} of Definition~\ref{def:simpsets}.

\begin{figure}[h]
  \centering
  \includegraphics[width=0.40\textwidth]{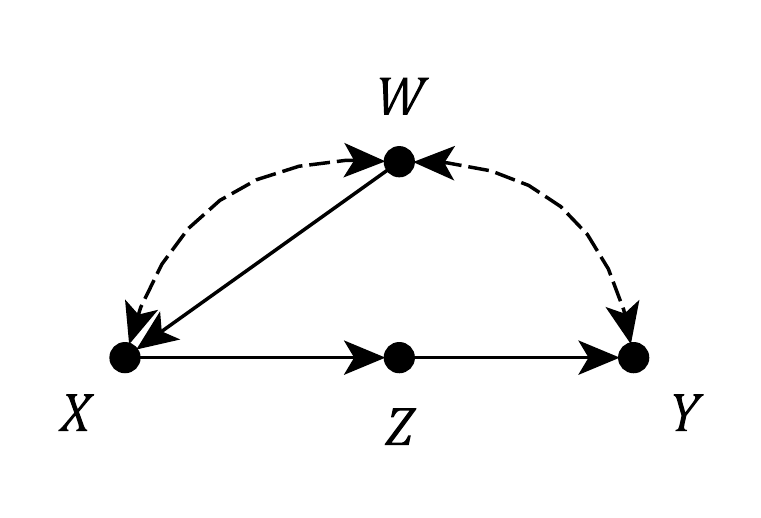}
  \caption{A graph $G$ for the example on the use of simplification sets.}
  \label{fig:simpsets}
\end{figure}

We show how simplification sets can be used in practice to derive a simpler expression via an example. We consider the causal effect of $\{X,Z,W\}$ on $Y$ in the graph $G$ of Figure~\ref{fig:simpsets}.

\noindent
The effect in question is identifiable and the ID algorithm readily gives atomic expression
\[  \sum_{X,W} P(Y|X,W,Z)P(X|W)P(W).  \]
We consider simplification sets with respect to $V_j = W$. The topological order is $W < X < Z < Y$. The atomic expression does not contain a term for $Z$ so we have $\+ M = \{Z\}$. By noting that $(Z \indep W | X)_G$ we are able to satisfy condition \eqref{eq:independence} of Definition~\ref{def:simpsets}. We can write
\[  P(Y,Z,X,W) = P(Z|X,W) P(Y|X,W,Z)P(X|W)P(W),  \]
as required by condition \eqref{eq:factorization} of Definition~\ref{def:simpsets} by setting $\+ E_Z = \{X,W\}$. Thus, the simplification sets $\+ D$ and $\+ E_Z$ for the atomic expression with respect to $W$ are $\emptyset$ and $\{X,W\}$, respectively. Finally, we obtain the simpler atomic expression by carrying out the summation over $W$:
\[  \sum_{X} P(Y|X,Z)P(X). \]

Neither Definition~\ref{def:simpsets} nor Theorem~\ref{thm:simpl} provide a method to obtain simplification sets or to determine whether they exist in general. To solve this problem we present a simplification algorithm for $\pi$-consistent atomic expressions that operates by constructing simplification sets iteratively for each variable in the summation set.

\begin{algorithm}[h]
  \begin{algorithmic}[1]
  \Function{simplify}{$A, G, \pi$}
  \State $j \gets 0$
  \While{$j < |\+ S|$}  \label{line:whileout}
    \State $B \gets A$ \label{line:backup}
    \State $\+ J \gets \emptyset$
    \State $\+ D \gets \emptyset$
    \State $\+ R \gets \emptyset$
    \State $\+ I \gets \emptyset$
    \State $j \gets j + 1$
    \State $i \gets \Call{index.of}{A,j}$  \label{line:indexof}
    \State $\+ M \gets \Call{get.missing}{A,G,j}$ \label{line:missing}
    \State $k \gets 1$
    \While{$k \leq i$} \label{line:whilein}
      \State $\langle \+ J_{\mathrm{new}}, \+ D_{\mathrm{new}}, \+ R_{\mathrm{new}} \rangle \gets \Call{join} {\+ J, \+ D, V_k, \+ C_k, S_j, \+ M, G, \pi} $ \label{line:join}
      \If{$ \+ J_{\mathrm{new}} \subseteq \+ J $} \label{line:failsimp}
        \State \textbf{break}
      \Else
        \State $\+ J \gets \+ J_{\mathrm{new}}$ \label{line:updateJ}
        \State $\+ D \gets  \+ D_{\mathrm{new}}$ \label{line:updateD}
        \If{$\+ R_{\mathrm{new}} \neq \emptyset$} \label{line:Rnotempty}
          \State $\+ R \gets \+ R \cup \+ R_{\mathrm{new}}$ \label{line:updateR}
          \State $\+ I \gets \+ I \cup \{ \+ D \}$ \label{line:updateI}
          \State $\+ M \gets \+ M \setminus \+ R_{\mathrm{new}}$ \label{line:updateM}
        \Else
          \State $k \gets k + 1$
        \EndIf
      \EndIf
    \EndWhile
    \If{$k = i+1$}
      \State $A_{\mathrm{new}} \gets \Call{factorize} {\+ J, \+ D,\+ R, \+ I, A}$ \label{line:factorize}
      \If{$A_{\mathrm{new}} = A$}
        \State $A \gets B$ \label{line:restore}
      \Else
        \State $A \gets A_{\mathrm{new}}$ \label{line:updateA}
        \State $\+S \gets \+S \setminus \{S_{j}\}$ \label{line:updateS}
        \State $j \gets 0$ \label{line:complete3}
      \EndIf
    \EndIf
  \EndWhile
  \State \textbf{return} {$A$}
  \EndFunction
  \end{algorithmic}
  \caption{Simplification of an atomic expression $A = \langle \+ T, \+ S \rangle$ given graph $G$ and topological ordering $\pi$.}
  \label{alg:simplify}
\end{algorithm}

Algorithm~\ref{alg:simplify} always attempts to perform maximal simplification, meaning that as many variables of the set $\+ S$ are removed as possible. If the simplification in terms of the entire set $\+ S$ can not be completed, the intermediate result with as many variables simplified as possible is returned. If simplification in terms of specific variables or a subset is preferred, the set $\+ S$ should be defined accordingly. 

The function $\Call{simplify}{}$ takes three arguments: an atomic expression $A[\+ W]$ that is to be simplified, a graph $G$ and a topological ordering $\pi$ of its vertices. $A$ is assumed to be $\pi$-consistent.

On line~\ref{line:indexof} the function $\Call{index.of}{}$ returns the corresponding index $i$ of the term containing $S_j$. Since $A$ is $\pi$-consistent, we only have to iterate through the variables $V_{1},\ldots,V_{j}$ as the terms outside this range contain no relevant information about the simplification of $V_j$.
The variables without a corresponding term in the atomic expression $A$ are retrieved on line~\ref{line:missing} by the function $\Call{get.missing}{}$. This function returns the set $\+ M$ of Definition~\ref{def:simpsets} with respect to the current variable to be summed over.

In order to show that the term of $A$ represent some joint distribution, we proceed in the order dictated by the topological ordering of the vertices. The sets $\+ J$ and $\+ D$ keep track of the variables that have been successfully processed and of the conditioning set of the joint term that was constructed on the previous iteration. Similarly, the sets $\+ R$ and $\+ I$ keep track of the variables and conditioning sets of the corresponding variables that the atomic expression does not originally contain a term for. Iteration through relevant terms begins on line~\ref{line:whilein}. Next, we take a closer look at the function $\Call{join}{}$ which is called next on line~\ref{line:join}.

\begin{algorithm}[h]
  \begin{algorithmic}[1]
  \Function{join}{$\+ J, \+ D, V, \+ C, S, \+ M, G, \pi$}
    \If{$\+ J = \emptyset$}
      \State \textbf{return} {$\langle \{V\}, \+ C, \emptyset  \rangle$} \label{line:initial}
    \EndIf
    \State $\+ G \gets \+ J^\pi \setminus An^*(V)_G$ \label{line:Gset}
    \State $\+ P \gets \mathcal{P}(\+ G)$ \label{line:Gpowerset}
    \State $n \gets |\+ P|$
    \For{$i = 1:n$} \label{line:forloop1}
      \State $\+ A \gets (An^*(V)_G \cup \+ P_i) \,\triangle\, \+ D$ \label{line:Aset}
      \State $\+ B \gets (An(V)_G \cup \+ P_i) \,\triangle\, \+ C $ \label{line:Bset}
      \If{$  (\+ J \indep \+ A | \+ D  \setminus \+ A)_G \textbf{ and } (V \indep \+ B| \+ C \setminus \+ B)_G   $} \label{line:checkcond}
        \State \textbf{return} {$\langle\+ J \cup \{V\}, (An(V)_G \cup \+ P_i), \emptyset  \rangle$}
      \EndIf
    \EndFor
    \If{$ \+M \neq \emptyset$} \label{line:Mnotempty}
      \For{$M^\prime \in \+ M$} \label{line:Miterate}
        \If{$M^\prime \in \+ D, M^\prime \not\in \+ C$} \label{line:checkM}
          \State $\langle \+ J_{\mathrm{new}}, \+ D_{\mathrm{new}}, \+ R \rangle \gets \Call{insert}{\+ J, \+ D, M^\prime, S, G, \pi}$
          \If{$ \+ J \subset \+ J_{\mathrm{new}}$}
            \State \textbf{return} {$\langle\+  J_{\mathrm{new}}, \+ D_{\mathrm{new}} , \+ R   \rangle$}
          \EndIf
        \EndIf
      \EndFor
    \EndIf
    \State \textbf{return} {$\langle \+ J, \+ D , \emptyset  \rangle$} \label{line:joinfail}
  \EndFunction
  \end{algorithmic}
  \caption{Construction of the joint distribution of the set $\+ J$ and a variable $V$ given their conditional sets $\+ D$ and $\+ C$ using d-separation criteria in $G$. $S$ is the current summation variable, $\+ M$ is the set of variables not contained in the expression and $\pi$ is a topological ordering.}
  \label{alg:join}
\end{algorithm}

Here $\mathcal{P}(\cdot)$ denotes the power set, $\triangle$ denotes the symmetric difference and $An^*(\cdot)_G$ denotes the ancestors with the argument included. The function $\Call{join}{}$ attempts to combine the joint term $P(\+ J| \+ D)$, obtained from the previous iteration steps, with the term $P(V| \+ C) := P(V_k|\+ C_k)$ of the current iteration step. d-separation statements of $G$ are evaluated to determine whether this can be done. In practice this means finding a suitable subset $\+ P_i$ of $\+ G$, where $\+ G \cup An(V)_G$ is the largest possible conditioning set of the new combined term. The set $\+ G$ is computed on line on line~\ref{line:Gset} of Algorithm~\ref{alg:join}. A valid subset $\+ P_i$ satisfies $P(\+ J| \+ D) = P(\+ J|An^*(V)_G, \+ P_i)$ and $P(V| \+ C) = P(V| An(V)_G, \+ P_i)$ which allow us to write the product $P(\+ J| \+ D)P(V| \+ C)$ as $P(\+ J, V| An(V)_G, \+ P_i)$. 

In order to find this valid subset, we compute the sets $\+ A$ and $\+ B$ for each candidate on lines~\ref{line:Aset} and \ref{line:Bset}. These sets characterize the necessary change in the conditioning sets of the terms $P(\+ J| \+ D)$ and $P(V| \+ C)$ that would enable a joint term to be formed by these two terms. The validity of the candidate set is finally checked on line~\ref{line:checkcond} which determines if the necessary change is allowed by d-separation criteria in the graph $G$. If no valid subset $\+ P_i$ can be found, we can still attempt to insert a missing variable of $\+ M$ by calling $\Call{insert}{}$. If this does not succeed either, the original sets $\+ J$ and $\+ D$ are returned, which instructs $\Call{simplify}{}$ to terminate simplification in terms of $V_j$ and attempt simplification in the next variable.

A special case where the first variable of the joint distribution forms $P(\+ J, \+ D)$ alone is processed on line 2 of Algorithm~\ref{alg:join}. In this case, we have an immediate result without having to iterate through the subsets of $\+ G$. The formulation of the set $\+ G$ ensures that the resulting factorization is $\pi$-consistent if it exists. Knowing that the ancestral set $An(V)_G$ has to be a subset of the new conditioning set also greatly reduces the amount of subsets we have to iterate through. In a typical situation, the size of $\+ P$ is not very large. Let us now inspect the insertion procedure in greater detail.

\begin{algorithm}[h]
  \begin{algorithmic}[1]
  \Function{insert}{$\+ J, \+ D, M^\prime, S, G, \pi$}
    \State $\+ G \gets \+ J^\pi \setminus An^*(M^\prime)_G$ \label{line:Gsetinsert}
    \State $n \gets |\+ G|$
    \For{$i = 1:n$}
      \State $\+ A \gets (An^*(M^\prime)_G \cup \+ P_i) \,\triangle\, \+ D $ \label{line:Asetinsert}
      \State $\+ B \gets (An(M^\prime)_G \cup \+ P_i) $ \label{line:Bsetinsert}
      \If{$  (\+ J \indep \+ A | \+ D  \setminus \+ A)_G \textbf{ and } (M^\prime \indep \+ S| \+ B \setminus \+ S)_G $} \label{line:checkindep}
        \State \textbf{return} {$\langle\+ J \cup \{M^\prime\},  (An^*(M^\prime)_G \cup \+ P_i), \{M^\prime\} \rangle$}
      \EndIf
    \EndFor
    \State \textbf{return} {$\langle \+ J, \+ D, \emptyset \rangle$} \label{line:insertfail}
  \EndFunction
  \end{algorithmic}
  \caption{Insertion of variable $M^\prime$ into the joint term $P(\+ J|\+ D)$ using d-separation criteria in $G$. $S$ is the current summation variable and $\pi$ is a topological ordering.}
  \label{alg:insert}
\end{algorithm}

In essence, the function $\Call{insert}{}$ is a simpler version of $\Call{join}{}$, because the only restriction on the conditioning set of $M^\prime$ is imposed by the conditioning set of $\+ J$ and the fact that $M^\prime$ has to be conditionally independent of the current variable $S$ to be summed over.
If $\Call{join}{}$ or $\Call{insert}{}$ was unsuccessful in forming a new joint distribution, we have that $\+ J_{\text{new}} \subset \+ J$. In this case simplification in terms of the current variable cannot be completed. If we have that $\+ J_{\text{new}} \not\subset \+ J$ the iteration continues.

Together the functions $\Call{join}{}$ and $\Call{insert}{}$ capture the two conditions of Definition~\ref{def:simpsets}. They are essentially two variations of the underlying procedure of determining whether the terms of the atomic expression actually represent a joint distribution. The only difference is that $\Call{join}{}$ is called when we are processing terms that already exist in the expression, and $\Call{insert}{}$ is called when there are variables without corresponding terms in the expression, that is the set $\+ M$ of Definition~\ref{def:simpsets} is not empty.

If the innermost while-loop of Algorithm~\ref{alg:simplify} succeeded in iterating through the relevant variables, we are ready to complete the simplification process in terms of $S_j$. We carry out the summation over $S_j$ which results in $P(\+ J \setminus \{V_i\}|\+ D)$. This is done on line~\ref{line:factorize} by calling $\Call{factorize}{\+ J, \+ D, \+ R, \+ I, A}$ which checks whether the joint term $P(\+ J \setminus \{V_i\}|\+ D)$ can be factorized back into a product of terms. In practice this means that if the function succeeds, it will return an atomic expression obtained by removing each inserted term $P(R|\+ I_R)$ such that $R \in \+ R$ and $\+ I_R \in \+I$ from atomic expression $A$. The status of the atomic expression is updated on lines~\ref{line:updateA} and \ref{line:updateS} to reflect this. If the function fails, it will return $A$ unchanged.

If the innermost while-loop did not iterate completely through the relevant variables, the simplification was not successful in terms of $S_j$ at this point. In this case we reset $A$ to its original state on line~\ref{line:restore} and attempt simplification in terms of the next variable. If there are no further variables to be eliminated, the outermost while-loop will also terminate. In the next theorem, we show that Algorithm~\ref{alg:simplify} is both sound and complete in terms of simplification sets. The proof for the theorem can be found in Appendix~\ref{app:complete}.

\begin{theorem} \label{thm:complete}  Let $G^\prime$ be a DAG and let $G$ be a subgraph  of $G^\prime$ over a vertex set $\+ W$ with a topological ordering $\pi$. Let $A[\+ W] = \langle \+ T, \{V_j\} \rangle$ be a $\pi$-consistent atomic expression. Then if $\Call{simplify}{A,G,\pi}$ succeeds, it has constructed a collection of simplification sets of $A$ with respect to $V_j$. Conversely, if there exists a collection of simplifications sets of $A$ with respect to $V_j$, then $\Call{simplify}{A,G,\pi}$ will succeed.
\end{theorem}

\section{High Level Algorithms} \label{sect:algo}
In this section, we present an algorithm to simplify all atomic expressions in the recursive stack of an expression. We will also provide a simple procedure to simplify quotients defined by two expressions: one representing the numerator and another representing the denominator. In some cases it is also possible to eliminate the denominator by subtracting common terms. First, we present a general algorithm to simplify topologically consistent expressions.

\begin{algorithm}[h]
  \begin{algorithmic}[1]
  \Function{deconstruct}{$B, G, \pi$}
  \State $\+ R \gets \emptyset$
  \For{$Y \in \+ A$}
    \State $\langle \{ \langle V_1, \+ C_1 \rangle, \ldots, \langle V_n, \+ C_n \rangle\},  \+ S_Y \rangle \gets \Call{simplify}{Y, G, \pi}$ \label{line:checkatomic}
    \If{$\+ S_Y = \emptyset$} 
      \State $\+ A \gets \+ A \cup \left( \bigcup_{i = 1}^n \{ \langle \{ \langle V_i, \+ C_i \rangle \}, \emptyset \rangle \} \right)$ \label{line:mergeatomic}
    \EndIf
  \EndFor
  \For{$ \langle \+ B_X, \+ A_X, \+ S_X \rangle \in \+ B$} 
    \State $\langle \+ B_X, \+ A_X, \+ S_X \rangle \gets \Call{deconstruct}{\langle \+ B_X, \+ A_X, \+ S_X \rangle, G}$ \label{line:bottom}
    \If{$\+ B_X = \emptyset \textbf{ and } \+ S_X = \emptyset$} \label{line:upone}
      \State $\+ R \gets \+ R \cup \{\langle \+ B_X, \+ A_X, \+ S_X \rangle\}$
      \State $\+ A \gets \+ A \cup \+ A_X$
    \EndIf
  \EndFor
  \State $\+ B \gets \+ B \setminus \+ R$
  \State \textbf{return} {$\langle \+ B, \+ A, \+ S \rangle$}
  \EndFunction
  \end{algorithmic}
  \caption{Recursive wrapper for the simplification of an expression $B = \langle \+ B, \+ A, \+ S \rangle$ given graph $G$ and topological ordering $\pi$.}
  \label{alg:deconstruct}
\end{algorithm}

Algorithm~\ref{alg:deconstruct} begins by simplifying all atomic expressions contained in the expressions. If an atomic expression contains no summations after the simplification but does contain multiple terms, each individual term is converted into an atomic expression of their own. After this, we iterate through all sub-expressions contained in the expression. The purpose of this is to carry out the simplification of every atomic expression in the stack and collect the results into as few atomic expressions as possible. First, we traverse to the bottom of the stack on line~\ref{line:bottom} by deconstructing sub-expressions until they have no sub-expressions of their own. Afterwards, it must be the case that $\langle \+ B_X, \+ A_X, \+ S_X \rangle$ consists of atomic sub-expressions only. 

If $\langle \+ B_X, \+ A_X, \+ S_X \rangle $ contains no summations on line~\ref{line:upone} then the atomic expressions contained in this expression do not require an additional expression to contain them, but can instead be transferred to be a part of the expression above the current one in the recursive stack. 
On line~\ref{line:mergeatomic} we lift the atomic expressions contained in the atomic sub-expressions up to the current recursion stage. 

There is no guarantee, that the resulting atomic expression is still $\pi$-consistent after this procedure. The function $\Call{deconstruct}{}$ operates on the principle of simplifying as many atomic expressions as possible, combining the results into new atomic expressions and simplifying them once more. We do not claim that this procedure is complete in a sense that Algorithm~\ref{alg:deconstruct} would always  find the simplest representation for a given expression. This method in nonetheless sound and finds drastically simpler expressions in almost every situation where such an expression exists.

We may also consider quotients often formed by deriving conditional distributions. For this purpose we need a subroutine to extract terms from atomic sub-expression that are independent of the summation index, that is $V_i \not\in \+ S$ and $\+ C_i \cap \+ S = \emptyset$.

\begin{algorithm}[h]
  \begin{algorithmic}[1]
  \Function{extract}{$B, G, \pi$}
    \State $B \gets \Call{deconstruct}{B, G, \pi}$ \label{line:ext:deconstruction}
    \If{$\+ S = \emptyset$} \label{line:extract:empty:sum}
      \For{$ X \in \+ B$} 
        \State $X \gets \Call{extract}{X , G, \pi}$
      \EndFor
      \For{$\langle \+ T_A, \+ S_A \rangle \in \+ A$} \label{line:extract:iterate:atomic}
        \If{$ \+S_A \neq \emptyset$}
          \State $\+ A_E \gets \emptyset$
          \State $\+ R \gets \emptyset$
          \For{$ \langle V, \+ C \rangle \in \+ T_A $} \label{line:extract:iterate:terms}
             \If{$ V \not\in \+ S_A \textbf{ and } \+ C \cap \+ S_A = \emptyset$} \label{line:extract:independent}
               \State $\+ A_E \gets \+ A_E \cup \{ \langle \{ \langle V, \+ C \rangle \} , \emptyset \rangle \}$
               \State $\+ R \gets \+ R \cup \{\langle V, \+ C \rangle \}$
             \EndIf
          \EndFor
          \State $\+ A \gets \+ A \cup \+ A_E$
          \State $\+ T_A \gets \+ T_A \setminus \+ R$
         \EndIf
      \EndFor
    \Else
      \State $\+ A_E \gets \emptyset$
      \State $\+ R \gets \emptyset$
      \For{$\langle \+ T_A, \+ S_A \rangle \in \+ A$}
        \If{$ \+S_A = \emptyset$}
          \State $\+ T_A^{(1)} \gets \emptyset$
          \State $\+ T_A^{(2)} \gets \emptyset$
          \For{$ \langle V, \+ C \rangle \in \+ T_A $} 
            \State $\+ T_A^{(1)} \gets \+ T_A^{(1)} \cup \{V\}$
            \State $\+ T_A^{(2)} \gets \+ T_A^{(2)} \cup \+ C$
          \EndFor
          \If{$\+ T_A^{(1)}\cap \+ S = \emptyset \textbf{ and } \+ T_A^{(2)} \cap \+ S = \emptyset$}
            \State $\+ A_E \gets \+ A_E \cup \{ \langle \+ T_A, \+ S_A \rangle \}$
            \State $\+ R \gets \+ R \cup \{ \langle \+ T_A, \+ S_A \rangle \}$
          \EndIf
        \EndIf
      \EndFor
      \State $\+ A \gets \+ A \setminus \+ R$
      \State $\+ B_E \gets \{B\}$
      \State \textbf{return} {$\langle \+ B_E, \+ A_E, \emptyset \rangle$}
    \EndIf
  \EndFunction
  \end{algorithmic}
  \caption{Extraction of terms independent of the summation indices from a expression $B = \langle \+ B, \+ A, \+ S \rangle$ given graph $G$ and topological ordering $\pi$.}
  \label{alg:extract}
\end{algorithm}

The procedure of Algorithm~\ref{alg:extract} is rather straightforward. First, we attempt to simplify $B$ by using $\Call{deconstruct}{}$ on line~\ref{line:ext:deconstruction}. Next, we simply recurse as deep as possible without encountering a sum in an expression. If a sum is encountered, extraction is attempted. On any stage where a sum was not encountered, we may still have atomic sub-expression that contain sums. Because the recursion had reached this far, we know that there are no summations above them in the stack, so we can attempt extraction on them as well. 

\begin{algorithm}[h]
  \begin{algorithmic}[1]
  \Function{$q$-simplify}{$B_1, B_2, G, \pi$}
  \State $B_1 \gets \Call{extract}{B_1, G, \pi}$ \label{line:ext1}
  \State $B_2 \gets \Call{extract}{B_2, G, \pi}$ \label{line:ext2}
  \If{$\+ S_1 \neq \emptyset \textbf{ or } \+ S_2 \neq \emptyset$} \label{line:sumempty}
    \State \textbf{return} {$ \langle B_1, B_2 \rangle $}
  \EndIf
  \State $i \gets 1$
  \While{$i \leq |\+ B_1| \textbf{ and } |\+ B_1| > 0 \textbf{ and } |\+ B_2| > 0$} \label{line:qsimp:iterateB}
    \For{$j = 1:|\+ B_2|$}
      \If{$B_{1i} = B_{2j}$}
        \State $\+ B_1 \gets \+ B_1 \setminus \{B_{1i}\}$
        \State $\+ B_2 \gets \+ B_2 \setminus \{B_{2j}\}$
        \State $i \gets 0$
        \State \textbf{break}
      \EndIf
    \EndFor
    \State $i \gets i + 1$
  \EndWhile
  \State $i \gets 1$
  \While{$i \leq |\+ A_1| \textbf{ and } |\+ A_1| > 0 \textbf{ and } |\+ A_2| > 0$} \label{line:qsimp:iterateA}
    \For{$j = 1:|\+ A_2|$}
      \If{$A_{1i} = A_{2j}$}
        \State $\+ A_1 \gets \+ A_1 \setminus \{A_{1i}\}$
        \State $\+ A_2 \gets \+ A_2 \setminus \{A_{2j}\}$
        \State $i \gets 0$
        \State \textbf{break}
      \EndIf
    \EndFor
    \State $i \gets i + 1$
  \EndWhile
  \State \textbf{return} {$ \langle B_1, B_2 \rangle $}
  \EndFunction
  \end{algorithmic}
  \caption{Simplification of a quotient $P_{B_1}/P_{B_2}$ given by the values of two expressions $B_1 = \langle \+ B_1, \+ A_1, \+ S_1 \rangle$ and $B_2 = \langle \+ B_2, \+ A_2, \+ S_2 \rangle$ given graph $G$ and topological ordering $\pi$.}
  \label{alg:simplifyquotient}
\end{algorithm}

Algorithm~\ref{alg:simplifyquotient} takes two expressions, $B_1$ and $B_2$, and removes any sub-expressions and atomic sub-expressions that are shared by $B_1$ and $B_2$. This is of course only feasible when the summation sets are empty for both $B_1$ and $B_2$. This condition is checked on line~\ref{line:sumempty}.

\section{Examples} \label{sect:exam}

\begin{figure}[h]
  \centering
  \includegraphics[width=0.32\textwidth]{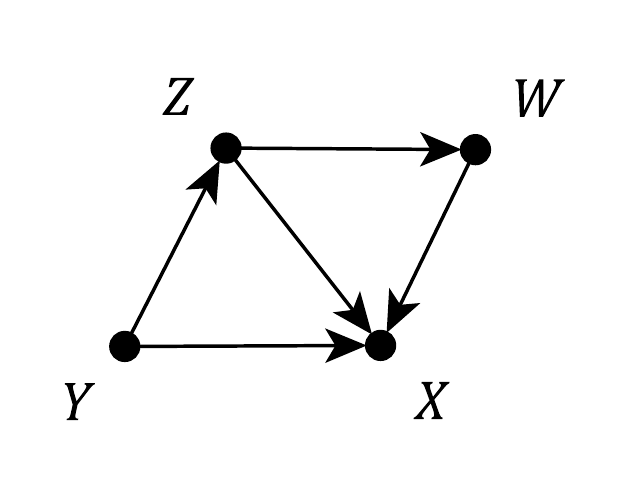}
  \caption{A graph $G$ for the example on the necessity of the insertion procedure.}
  \label{fig:ginsert}
\end{figure}

In this section we present examples of applying the algorithms of the previous sections. We denote line number $y$ of algorithm $x$ with A$x$:$y$. We begin with a simple example on the necessity of the $\Call{insert}{}$ procedure in graph $G$ of Figure~\ref{fig:ginsert}.

The causal effect of $W$ on $X$ is identifiable in this graph, and expression
\[  \sum_{Z,Y} P(Y)P(Z|Y)P(X|W,Z,Y) \]
is obtained by direct application of the ID algorithm or by the truncated factorization formula for causal effects in Markovian models \citep{pearl09}.
We let $A$ be this atomic expression. The topological ordering $\pi$ is $X > W > Z > Y$ and $\+ M = \{W\}$. The call to $\Call{simplify}{A,G,\pi}$ will first attempt simplification in terms of $Z$, by calling 
\[ \Call{join}{\emptyset, \emptyset, X, \{W,Z,Y\}, Z, \{W\}, G, \pi}, \] 
which results in $\langle X, \{W,Z,Y\}, \emptyset \rangle$. At the second call 
\[ \Call{join}{\{X\}, \{W,Z,Y\}, Z, Y, Z, \{W\}, G, \pi} \] 
we already run into trouble since we cannot find a conditioning set that would allow $Z$ to be joined with $\{X\}$. However, since $\+ M$ is non-empty and $W \in \{W,Z,Y\}$ and $W \not\in \{Z\}$ this means that the next call is
\[ \Call{insert}{\{X\}, \{W,Z,Y\}, W, Z, G, \pi}. \]
Insertion fails in this case, as one can see from the fact that no conditioning set exists that would make $W$ conditionally independent of $Z$. Thus we recurse back to $\Call{join}{}$ and back to $\Call{simplify}{}$ and end up on line~\lineref{alg:simplify}{line:failsimp} which breaks out of the while-loop. Thus  $A$ cannot be simplified in terms of $Z$. Simplification is attempted next in terms of $Y$. The first two calls are in this case
\[ \Call{join}{\emptyset, \emptyset, X, \{W,Z,Y\}, Y, \{W\}, G, \pi}, \]
\[ \Call{join}{\{X\}, \{W,Z,Y\}, Z, \{Y\}, Y, \{W\}, G, \pi}, \]
and in the second call we run into trouble again and have to attempt insertion 
\[ \Call{insert}{\{X\}, \{W,Z,Y\}, W, Y, G, \pi}. \]
This time we find that we can add a term for $W$ which is $P(W|Z,Y)$ because $(W \indep Y|Z)_G$. The other calls to $\Call{join}{}$ also succeed and we can write the value of $A$ as
\[  \frac{\sum_{Z,Y} P(Y)P(Z|Y)P(W|Y,Z)P(X|W,Z,Y)}{P(W|Z)}. \]
and complete the summation in terms of $Y$. After the call to $\Call{factorize}{}$ we are left with the final expression
\[  \sum_{Z} P(X|W,Z)P(Z).  \]

We continue by considering again graph $G$ depicted in Figure~\ref{fig:grfG}. The topological ordering $\pi$ is $Y > Z_1 > Z_3 > X > Z_2$. Atomic expression $A_1$ given by
\[  \sum_{X,Y} P(Y|Z_2,X,Z_3,Z_1)P(Z_3|Z_2,X)P(X|Z_2)P(Z_2) , \]
is a part of the expression to be simplified.

We will first simplify $A_1$ and take a closer look at how the function $\Call{join}{}$ operates. The call to $\Call{simplify}{A_1, G, \pi}$ will attempt simplification in terms of the set $\{X,Y\}$ in the ordering that agrees with the topological ordering $\pi$, which is $(Y,X)$. After initializing the required sets, we find the index of the term with $Y$ as a variable on line~\ref{line:indexof}. There is one missing variable, $Z_1$, so $\+ M = \{Z_1\}$ as returned by $\Call{get.missing}{}$ on line~\lineref{alg:simplify}{line:missing}. The first call to $\Call{join}{}$ results in $\langle Y, \{Z_2,X,Z_3,Z_1\}, \emptyset \rangle$, because line~\lineref{alg:join}{line:initial} is triggered. Condition on line~\lineref{alg:simplify}{line:failsimp} is not satisfied since $\+ J_{\text{new}} = \{Y\} \not\subseteq \emptyset = \+ J$. Thus we update the status of $\+ J$ and $\+ D$ on lines~\lineref{alg:simplify}{line:updateJ} and \lineref{alg:simplify}{line:updateD}. Since $\+ R_{\text{new}} = \emptyset$ on line~\lineref{alg:simplify}{line:Rnotempty} we do not have to update the status of $\+ R, \+ I$ and $\+ M$ on lines~\lineref{alg:simplify}{line:updateR}, \lineref{alg:simplify}{line:updateI} and \lineref{alg:simplify}{line:updateM}. The innermost while-loop is now complete and we call $\Call{factorize}{}$ on line~\lineref{alg:simplify}{line:factorize} which succeeds in removing the term $P(Y|Z_2,X,Z_3,Z_1)$ by completing the sum. Now we update the status of the atomic expression on line~\lineref{alg:simplify}{line:updateA} and remove $Y$ from the set of variables to be summed over on line~\lineref{alg:simplify}{line:updateS}. The resulting value of the expression at this point is
\[  \sum_{X} P(Z_3|Z_2,X)P(X|Z_2)P(Z_2). \]
Next, the summation in terms of $X$ is attempted. $\Call{join}{}$ is once again successful, because $Z_3$ is the first variable to be joined and line~\lineref{alg:join}{line:initial} is triggered. Next we attempt to join the terms $P(Z_3|Z_2,X)$ and $P(X|Z_2)$. Computation of the set $\+ G$ on line~\lineref{alg:join}{line:Gset} results in
\[  \{Z_3\}^\pi \setminus An^*(X)_G = \{X, Z_2\} \setminus \{X, Z_2\} = \emptyset. \]
The power set computed on line~\lineref{alg:join}{line:Gpowerset} contains only the empty set. For $\+ P_1 = \emptyset$ we have
\[  A = (An(X)^*_G \cup \+ P_1) \triangle \+ D = (\{X, Z_2\} \cup \emptyset) \triangle \{X,Z_2\} = \emptyset  \]
on line~\ref{line:Aset}, and
\[  B = (An(X)_G \cup \+ P_1) \triangle \+ C = (\{Z_2\} \cup \emptyset) \triangle \{Z_2\} = \emptyset  \]
on line~\ref{line:Bset}. The condition on line~\lineref{alg:join}{line:checkcond} evaluates to true and we return with $\langle \{Z_3, X\}, \{Z_2\}, \emptyset \rangle$.
The innermost while-loop terminates allowing the summation over $X$ to be performed. The function $\Call{factorize}{}$ provides us with the final expression 
\begin{equation} \label{eq:simpxmp}
P(Z_3|Z_2)P(Z_2).
\end{equation}

Next, we will consider the full example and see how $\Call{$q$-simplify}{}$ is applied. Using the ID algorithm we obtain the causal effect of $X$ on $Z_1, Z_2, Z_3$ and $Y$ in graph $G$ of Figure~\ref{fig:grfG} and it is
\begin{align*}
 P(Z_1| Z_2,X)P(Z_3| Z_2) \frac{\sum_{X}P(Y| Z_2,X,Z_3,Z_1)P(Z_3| Z_2,X)P(X| Z_2)P(Z_2)}{\sum_{X,Y}P(Y| Z_2,X,Z_3,Z_1)P(Z_3| Z_2,X)P(X| Z_2)P(Z_2)} \times \\ \sum_{X,Z_3,Y}P(Y| Z_2,X,Z_3,Z_1)P(Z_3| Z_2,X)P(X| Z_2)P(Z_2).
\end{align*}
We will represent this as a quotient of expression using Definition~\ref{def:expr}.
Let $A_1$ be the atomic expression of the previous example and let $A_2$ also be an atomic expression given by
\[  \sum_{X} P(Y|Z_2,X,Z_3,Z_1)P(Z_3|Z_2,X)P(X|Z_2)P(Z_2) , \]
which is essentially the same as $A_1$, but with the variable $Y$ removed from the summation set $\+ S$. Similarly, we let $A_3$ be an atomic expression given by
\[  \sum_{X,Z_3,Y} P(Y| Z_2,X,Z_3,Z_1)P(Z_3| Z_2,X)P(X| Z_2)P(Z_2). \]
We also define the atomic expressions $A_4$ with the value $P(Z_3|Z_2)$ and $A_5$ with the value $P(Z_1|Z_2,X)$.
Now, we define two expressions $B_1$ and $B_2$ for the quotient $P_{B_1}/P_{B_2}$ as follows:
\[  B_1 = \langle \emptyset, \{A_2,A_3, A_4, A_5\}, \emptyset \rangle, \quad B_2 = \langle \emptyset, \{A_1\}, \emptyset \rangle. \]
We now call $\Call{$q$-simplify}{B_1,B_2,G,\pi}$. First, we must trace the calls to $\Call{extract}{}$ for both expressions on lines~\lineref{alg:simplifyquotient}{line:ext1} and \lineref{alg:simplifyquotient}{line:ext2}. For $B_1$ and $B_2$ this immediately results in a call to $\Call{deconstruct}{}$ on line~\lineref{alg:extract}{line:ext:deconstruction}. First, the function applies $\Call{simplify}{}$ to each atomic expression contained in the expressions on line~\lineref{alg:deconstruct}{line:checkatomic}.

Let us first consider the simplification of $A_2$. As before with $A_1$, we have that $\Call{join}{}$ first succeeds in forming $\langle Y, \{Z_2,X,Z_3,Z_1\}, \emptyset \rangle$, but this time $Y$ is not in the summation set, so we continue. Next, the algorithm attempts to join $P(Y|Z_2,X,Z_3,Z_1)$ with $P(Z_3|Z_2,X)$. The set $\+ G$ is defined as
\[  \{Y\}^\pi \setminus An^*(Z_3)_G = \{Z_3,Z_1,X,Z_2\} \setminus \{Z_3,Z_2\} = \{Z_1, X\}  \]
and its subsets are $\{Z_1, X\}$, $\{Z_1\}$, $\{X\}$ and $\emptyset$. For the first subset $\+ P_1 = \emptyset$ we have that

\[  A = (An^*(Z_3) \cup \+ P_1) \triangle \+ D = \{Z_2, Z_3\} \triangle \{Z_2,X,Z_3,Z_1\} = \{X,Z_1\}  \] and since $(Y \not\indep X,Z_1|Z_3,Z_2)_G$ the condition on line~\lineref{alg:join}{line:checkcond} is not satisfied. We continue with $\+ P_2 = \{X\}$ and obtain
\[  A = (An^*(Z_3) \cup \+ P_2) \triangle \+ D = \{X,Z_2, Z_3\} \triangle \{Z_2,X,Z_3,Z_1\} = \{Z_1\}  \] and since $(Y \not\indep Z_1|X, Z_3,Z_2)_G$ the condition on line~\lineref{alg:join}{line:checkcond} is still not satisfied. Next, for $\+ P_3 = \{Z_1\}$ we have
\[  A = (An^*(Z_3) \cup \+ P_3) \triangle \+ D = \{Z_2, Z_3, Z_1\} \triangle \{Z_2,X,Z_3,Z_1\} = \{X\}  \] and since $(Y \not\indep X|Z_1, Z_3,Z_2)_G$ the condition on line~\lineref{alg:join}{line:checkcond} is again, not satisfied. Finally, for $\+ P_4 = \{Z_1, X\}$ we have
\[  A = (An^*(Z_3) \cup \+ P_4) \triangle \+ D = \{Z_2,X,Z_3,Z_1\} \triangle \{Z_2,X,Z_3,Z_1\} = \{X\}  \] and
\[  B = (An(Z_3) \cup \+ P_4) \triangle \+ C = \{Z_2,X,Z_1\} \triangle \{Z_2,X\} = \{Z_1\}. \] Both conditions on line~\lineref{alg:join}{line:checkcond} are now satisfied ny noting that $(Z_3 \indep Z_1|X,Z_2)_G$. Afterwards we obtain 
\[ P(Y|Z_2,X,Z_3,Z_1)P(Z_3|Z_2,X) = P(Y,Z_3|Z_1,Z_2,X) \] 
and continue in an attempt to join the term $P(X|Z_2)$ with this result. The set $\+ G$ is now defined as 
\[  \{Y,Z_3\}^\pi \setminus An^*(X)_G = \{Z_1,X,Z_2\} \setminus \{X, Z_2\} = \{Z_1\}  \]
and its subsets are $\{Z_1\}$ and $\emptyset$. Starting with $\+ P_1 = \emptyset$ we have that 
\[  A = (An^*(X) \cup \+ P_1) \triangle \+ D = \{X, Z_2\} \triangle \{Z_1,Z_2,X\} = \{Z_1\}  \] and since $(Y,Z_3 \not\indep Z_1|X,Z_2)_G$ the condition on line~\lineref{alg:join}{line:checkcond} is not satisfied. Continuing with $\+ P_2 = \{Z_1\}$ we have
\[  B = (An(X) \cup \+ P_2) \triangle \+ C = \{Z_2,Z_1\} \triangle \{Z_2\} = \{Z_1\}. \] Again, the condition on line~\lineref{alg:join}{line:checkcond} is not satisfied by noting that
$(X \not\indep Z_1 | Z_2)_G$. We have exhausted the possible subsets, which means that we enter the loop on line~\lineref{alg:join}{line:Miterate} since the set $\+ M = \{Z_1\}$ is not empty of line~\lineref{alg:join}{line:Mnotempty}.

In this case $\Call{insert}{}$ is called to bring $Z_1$ into the expression because $Z_1 \in \+ D = \{Z_1,Z_2,X\}$ and $Z_1 \not\in \+ C = \{Z_2\}$. The set $\+ G$ is constructed on line~\lineref{alg:insert}{line:Gsetinsert} and it is
\[  \+ J^\pi \setminus An^*(Z_1)_G = \{Y,Z_3\}^\pi \setminus \{X,Z_1,Z_2\} = \emptyset. \] For the only subset $\+ P_1 = \emptyset$ we have 
\[  B = (An(Z_1)_G \cup \+ P_1) = \{X, Z_2\}  \]
on line~\lineref{alg:insert}{line:Bsetinsert}, and since $(Z_1 \not\indep X| Z_2)_G$ the condition on line~\lineref{alg:insert}{line:checkindep} is not satisfied and we return with $\langle \+ J, \+ D, \emptyset \rangle$ unchanged on line~\ref{line:insertfail} of Algorithm~\ref{alg:insert}, which causes $\Call{join}{}$ to also return with the same output on line~\lineref{alg:insert}{line:joinfail}. The condition on line~\lineref{alg:simplify}{line:failsimp} is now satisfied and we cannot simplify $A_2$.

The atomic expression $A_3$ can be simplified. First, $Y$ is eliminated exactly as it was removed from $A_1$. Following the same principle we can see that whenever a variable in the summation set is the largest one in the topological order of the variables contained in the atomic expression, it will be removed successfully. From this we obtain that the value of $A_3$ is in fact simply $P(Z_2)$. Let us call the atomic expression with this value $E$, that is $P_E = P(Z_2)$. The atomic expression $A_1$ can also be simplified, and its value is given by (\ref{eq:simpxmp}). Furthermore, since this value is made of two product terms, it is split into two atomic expressions respectively. Let these be called $D_1$ and $D_2$ such that $P_{D_1} = P(Z_3|Z_2)$ and $P_{D_2} = P(Z_2)$. 

Applying $\Call{simplify}{}$ to $A_4$ and $A_5$ simply returns the original expressions, since they do not contain any summations and the loop on line~\lineref{alg:simplify}{line:whileout} is never entered. The set of atomic expressions is afterwards updated on line~\lineref{alg:deconstruct}{line:mergeatomic}. Neither $B_1$ nor $B_2$ contain any sub-expressions or summations on line~\lineref{alg:deconstruct}{line:upone}, so $\Call{deconstruct}{B_1,G,\pi}$ returns $\langle \emptyset, \{A_2,E,A_4,A_5\}, \emptyset \rangle$ and $\Call{deconstruct}{B_2,G,\pi}$ returns $\langle \emptyset, \{D_1,D_2\}, \emptyset \rangle$. The lack of summations on line~\lineref{alg:extract}{line:extract:empty:sum} of causes $\Call{extract}{}$ to iterate through the atomic expression contained in $B_1$ and $B_2$ directly on line~\lineref{alg:extract}{line:extract:iterate:atomic}, since neither of them have any sub-expressions of their own.

Only $A_2$ contains a sum at this point. The iteration over the terms of $A_2$ on line~\lineref{alg:extract}{line:extract:iterate:terms} finds that the only term that does not contain $X$ is $P(Z_2)$ on line~\lineref{alg:extract}{line:extract:independent}. Let us denote the atomic expression with the value $P(Z_2)$ as $C_1$ and the atomic expression resulting from the extraction as $C_2$ which now has the value
\[  \sum_{X} P(Y|Z_2,X,Z_3,Z_1)P(Z_3|Z_2,X)P(X|Z_2). \]
This completes the extraction and results in an expression $B_1'$ such that 
\[ B_1' = \langle \emptyset, \{C_1, C_2, E, A_4, A_5\}, \emptyset \rangle. \] 
The expression $B_2$ remains unchanged.

\Call{$q$-simplify}{} is now able to proceed. Neither $B_1'$ nor $B_2$ contain sub-expression so the loop on line~\lineref{alg:simplifyquotient}{line:qsimp:iterateB} is not entered, and we are only subtracting their common atomic expressions in the loop on line~\lineref{alg:simplifyquotient}{line:qsimp:iterateA}. It is easy to see that $A_4 = D_1$ and $C_1 = D_2$, so they are removed from both $B_1'$ and $B_2$.
Finally, the expressions corresponding to the numerator and denominator are returned.

To summarize, we began with the expression
\begin{align*}
 P(Z_1| Z_2,X)P(Z_3| Z_2) \frac{\sum_{X}P(Y| Z_2,X,Z_3,Z_1)P(Z_3| Z_2,X)P(X| Z_2)P(Z_2)}{\sum_{X,Y}P(Y| Z_2,X,Z_3,Z_1)P(Z_3| Z_2,X)P(X| Z_2)P(Z_2)} \times \\ \sum_{X,Z_3,Y}P(Y| Z_2,X,Z_3,Z_1)P(Z_3| Z_2,X)P(X| Z_2)P(Z_2).
\end{align*}
and successfully simplified it into
\[  P(Z_1|Z_2,X)P(Z_2) \sum_{X} P(Y|Z_2,X,Z_3,Z_1)P(Z_3|Z_2,X)P(X|Z_2). \]

\section{Discussion} \label{sect:disc}

We have presented a formal definition of topologically consistent atomic expressions and simplification sets and provided a sound and complete algorithm to find these sets for a given expression. We also discussed some general techniques that apply to a more general class of these expressions. Algorithm~\ref{alg:identify} and Algorithm~\ref{alg:condidentify}, presented in Appendix~\ref{app:consistent}, have been previously implemented in the R package causaleffect \citep{tikka17}. We have updated the package to include all of the simplification procedures presented in this paper and they can be applied to all causal effect and conditional causal effect expressions derived from identification procedures. Our definition of topologically consistent atomic expressions is similar to g-functionals that can be used to characterize identifiability results under special conditions \citep{shpitser16}.

It is plausible that these procedures could also be extended into other causal inference results, such as formulas for $z$-identifiability, transportability and meta-transportability of causal effects. The extensions are non-trivial however, since transportability formulas contain terms with distributions from multiple domains and $z$-identifiable causal effects contain do-operators in the conditioning sets which would require the implementation of the rules of do-calculus into Algorithm~\ref{alg:simplify}. Do-calculus consists of three inference rules that can be used to manipulate probabilities involving the do-operator \citep{pearl09}. Currently, we operate only on expressions that do not involve the do-operator. In fact, in our procedure it is not required to know the original causal query that produced the result. 

Simpler expressions have many useful properties. They can help in understanding and communicating results and evaluating them saves computational resources. Estimation accuracy can also be improved in some cases when variables that are present in the original expression suffer from missing data or measurement error. One example where the benefits of simplification are realized can be found in \citep{hyttinen2015}, where expressions of causal effects are derived and repeatedly evaluated for a large number of causal models. 

Our approach to simplification stems from the nature of causal effect expressions. In our setting, a question still remains whether simplification sets completely characterize all situations where a variable can be eliminated from an atomic expression. One might also consider simplification in a general setting, where we do not assume topological consistency or any other constraints for the atomic expressions. In this case a 'black box' definition for simplification could be considered, where we simply require that when the sum over a variable of interest is completed we are again left with another atomic expression without this variable in the summation set. This framework is theoretically interesting but we are not aware of any potential applications.

The worst case time complexity of Algorithm~\ref{alg:simplify} is difficult to gauge and is a topic for further research. One can observe that the performance of the algorithm is highly dependent on the size of the differences of the conditioning sets between adjacent terms. Both Algorithm~\ref{alg:join} and Algorithm~\ref{alg:insert} iterate through the subsets of these differences and check d-separation criteria for each subset. Thus dynamic programming solutions could be implemented to further improve performance by collecting the results of these checks. Previously determined conditional independences would not need to be checked again and could be retrieved from memory instead.

\begin{figure}[t]
  \centering
  \includegraphics[width=0.25\textwidth]{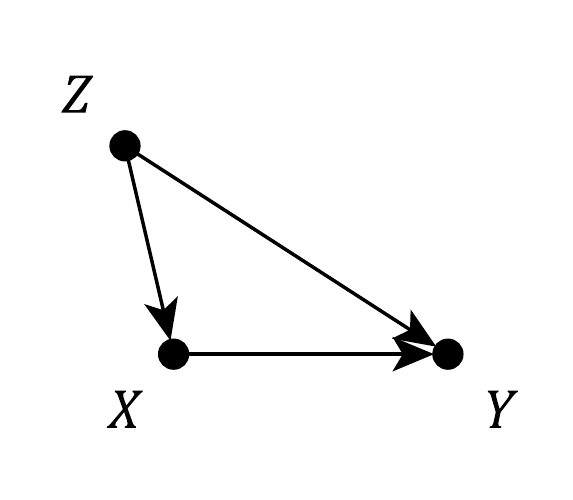}
  \caption{A graph $G$ for a situation where simplification fails}
  \label{fig:simpiden}
\end{figure}

In some cases, simplification has some apparent connections to identifiability. Consider the graph $G$ of Figure~\ref{fig:simpiden}.
\noindent
In this graph the causal effect of $X$ on $Y$ is identifiable, and its expression is \[ \sum_{Z}P(Y|Z,X)P(Z). \] If we let $Z$ be an unobserved variable instead, then $G$ depicts the well-known bow-arc graph, where the same causal effect is unidentifiable. This corresponds to an unsuccessful attempt to remove $Z$ from the expression of the causal effect. However, we cannot know beforehand whether an expression for a causal effect is going to be atomic or not, so we cannot use our algorithm to derive identifiability in general.

A reviewer suggested a simplification algorithm where the ID algorithm would be applied to latent projections \citep{pearl91} onto the variables to be marginalized. This algorithm would be able to solve many, but not all, simplification tasks. Importantly, in the example presented in Figure~\ref{fig:grfG}, we cannot make any variables latent, as we are interested in the causal effect of $X$ on all of the other variables. A reviewer also suggested that simplified expressions could be categorized into those that are obtained through latent projections and those that are not. This categorization might give additional insight into the topic.

\acks{We thank the anonymous reviewers for their constructive feedback that greatly helped us to improve this paper. We also thank Lasse Leskel\"{a} for his comments.}

\appendix
\section{Topological Consistency of Causal Effect Formulas}
\label{app:consistent}
We prove the statement that every causal effect formula returned by the algorithms of \citet{shpitser06,shpitser06cond} can be represented by using products and quotients of $\pi$-consistent atomic expressions such that $\pi$ is a topological ordering of $G$. 

We use the notation $G[\+ X]$ to denote an induced subgraph, which is obtained from $G$ by removing all vertices not in $\+ X$ and by keeping all edges between the vertices of $\+ X$ in $G$. Here $G_{\thickbar {\+ X},\thickubar {\+ Z}}$ means the graph that is obtained from $G$ by removing all incoming edges of $\+ X$ and all outgoing edges of $\+ Z$. We say that $G$ is an \emph{I-map} of $P$ if $P$ admits the causal Markov factorization with respect to $G = \langle \+ V, \+ E \rangle $, which is
\[  P = \prod_{i=1}^n P(V_i\vert Pa^*(V_i)_G) \prod_{j=1}^k P(U_j), \]
where $Pa^*(\cdot)$ contains unobserved parents as well.

Consider first lines 2, 3, 4 and 7 of Algorithm~\ref{alg:identify} where recursive calls occur and let $\pi_r$ be the topological ordering of the graph in the previous recursion step. Line 2 limits the identification procedure to the ancestors of $\+ Y$ so we can still obtain an expression that topologically consistent with respect to a topological ordering obtained from $\pi_r$ by removing non-ancestors. Lines 3 and 4 make no changes to the distribution $P$ and the graph $G$. On line 7 the induced subgraph $G[\+ S^\prime]$ in the next call is a C-component, but the joint distribution in this case is a $\pi_r$-consistent expression
\[ P(\+ S^\prime) = \prod_{V_i \in \+ S^\prime}P(V_i \vert V_i^{\pi_r} \cap \+ S^\prime,v_i^{\pi_r}\setminus \+ s^\prime), \]
 since every conditioning set is of the form $V_i^{\pi_r}$ when we only consider variables instead of their values, so we obtain
\[ P(\+ S^\prime) = \prod_{V_i \in \+ S^\prime}P(V_i | V_i^{\pi_r}), \]
Furthermore, any expression returned from line 7 will now be ${\pi_r}$-consistent. Thus all recursive calls retain topological consistency with respect to $\pi$.

Consider now the non-recursive terminating calls on lines 1 and 6. Consider line 1 first. If line two was triggered previously, we can factorize $P(\+ V)$ in such a way that each variable is conditioned by its ancestors, since the ancestors of ancestors of $\+ Y$ are by definition ancestors of $\+ Y$. If line 7 was triggered previously we already know that the joint distribution was previously factorized in a $\pi_r$-consistent fashion. If line 3 or 4 was triggered previously, we know that they have not imposed any changes on $P$ of $G$. Line 6 clearly produces a $\pi_r$ consistent end result. Lines 4, 6 and 7 can only produce either products of quotients. By noting that $\pi_r$-consistency implies $\pi$-consistency, we have that the result of the algorithm can always be represented by using products and quotients of $\pi$-consistent atomic expressions.

\begin{algorithm}[h]
	\begin{algorithmic}[1]
		\INPUT{Value assignments $\+ x$ and $\+ y$, joint distribution $P(\+ v)$ and a DAG $G = \langle \+ V, \+ E \rangle$. $G$ is an $I$-map of $P$.}
		\OUTPUT{Expression for $P_{\+ x}(\+ y)$ in terms of $P(\+ v)$ or \textbf{FAIL}$(F,F^\prime)$.}
		\Statex
		\Statex \textbf{function}{ \textbf{ID}$(\+ y, \+ x, P, G)$}
			\If{$\+ x = \emptyset$,}
				\Statex \textbf{\quad return}{ $\sum_{v \in \+ v \setminus \+ y}P(\+ v)$.}
			\EndIf
			\If{$\+ V \neq An(\+ Y)_G,$}
				\Statex \textbf{\quad return}{ \textbf{ID}$(\+ y, \+ x \cap An(\+ Y)_G, P(An(\+ Y)_G), G[An(\+ Y)_G)]$.}
			\EndIf	
			\State \textbf{Let}{ $\+ W = (\+ V \setminus \+ X) \setminus An(\+ Y)_{G_{\thickbar {\+ X}}}$.}
			\Statex \textbf{if}{ $\+ W \neq \emptyset $,} \textbf{then}
				\Statex \quad \textbf{return}{ \textbf{ID}$(\+ y, \+ x \cup \+ w, P, G)$.}
			\If{$C(G[\+ V \setminus \+ X]) = \{G[\+ S_1], \ldots,G[\+ S_k]\}$,}
				\Statex \textbf{\quad return}{ $\sum_{v \in \+ v \setminus (\+ y \cup \+ x)} \prod_{i=1}^k$ \textbf{ID}($\+ s_i, \+ v \setminus \+ s_i, P, G)$.}
			\EndIf
			\Statex \textbf{if}{ $C(G[\+V \setminus \+ X]) = \{G[\+ S]\}$,} \textbf{then}
				\State \quad \textbf{if}{ $C(G) = \{G\}$,} \textbf{then}
				\Statex \quad \quad \textbf{throw FAIL}{$(G, G[\+ S])$.}
				\State \quad \textbf{if}{ $G[\+ S] \in C(G)$,} \textbf{then} 
				\Statex \quad \quad \textbf{return} {$\sum_{v \in \+ s \setminus \+ y} \prod_{V_i \in \+ S}{P(v_i \vert v_i^\pi)}$.}
				\State \quad \textbf{if}{ $(\exists \+ S^\prime)\+ S \subset \+ S^\prime \text{ such that } G[\+ S^\prime] \in C(G)$,} \textbf{then}
				\Statex \quad \quad \textbf{return} {\textbf{ID}$(\+ y, \+ x \cap \+ s^\prime, \prod_{V_i \in \+ S^\prime}{P(V_i \vert V_i^\pi \cap \+ S^\prime,v_i^\pi \setminus \+ s^\prime), G[\+ S^\prime}])$.}
	\end{algorithmic}
	\caption{The causal effect of intervention $do(\+ X = \+ x)$ on $\+ Y$ \citep{shpitser06}.}
	\label{alg:identify}
\end{algorithm}

The claim is now apparent for Algorithm~\ref{alg:condidentify} since line~\ref{line:identifyrec} is eventually called for every conditional causal effect.

\begin{algorithm}[h]
	\begin{algorithmic}[1]
		\INPUT{Value assignments $\+ x$, $\+ y$ and $\+ z$, joint distribution $P(\+ v)$ and a DAG $G = \langle \+ V, \+ E \rangle$. $G$ is an $I$-map of $P$.}
		\OUTPUT{Expression for $P_{\+ x}(\+ y| \+ z)$ in terms of $P(\+ v)$ or \textbf{FAIL}$(F,F^\prime)$.}
		\Statex
		\Statex \textbf{function}{ \textbf{IDC}$(\+ y, \+ x,\+ z, P, G)$}
			\If{$\exists Z \in \+ Z \text{ such that } (\+ Y \indep Z|\+ X, \+ Z \setminus \{Z\})_{G_{\thickbar {\+ X},\thickubar {\+ Z}}}$}
				\Statex \textbf{\quad return} { \textbf{IDC}$(\+ y, \+ x \cup \{z\}, \+ z \setminus \{z\}, P, G)$.}
			\EndIf
			\State \textbf{else let} { $P^\prime =\;$\textbf{ID}$(\+ y \cup \+ z, \+ x, P, G)$.}
			\Statex \textbf{return} {$P^\prime / \sum_{y \in \+ y}P^\prime$} \label{line:identifyrec}
	\end{algorithmic}
	\caption{The causal effect of intervention $do(\+ X = \+ x)$ on $\+ Y$ given $\+ Z$ \citep{shpitser06cond}.}
	\label{alg:condidentify}
\end{algorithm}

\section{Proof of Theorem~\ref{thm:simpl}}
\label{app:simpl}

\begin{proof} By direct calculation we obtain
\begin{align*}
    P_{A} &= \sum_{V_j} \prod_{i = 1}^n P(V_i|\+ C_i) \\
&= \prod_{V_i < V_j} P(V_i|\+ C_i) \sum_{V_j} \prod_{V_i \geq V_j} P(V_i|\+ C_i) \\
&= \prod_{V_i < V_j} P(V_i|\+ C_i) \sum_{V_j} \frac{P(V_{\pi(p)},\ldots, V_{\pi(q)}|\+ D)}{\prod_{U \in \+ M} P(U | \+ E_U)} \\
&= \prod_{V_i < V_j} P(V_i|\+ C_i) \frac{P(V_{\pi(p+1)},\ldots, V_{\pi(q)}|\+ D)}{\prod_{U \in \+ M} P(U | \+ E_U)} \\
&= \prod_{V_i < V_j} P(V_i|\+ C_i) \prod_{V_i > V_j} P(V_i|\+ D_i) := P_{A^\prime},
\end{align*}
where the sets $\+ D_i$ are obtained from the factorization of the joint term such that $A^\prime$ is a $\pi^*$-consistent where $\pi^*$ is obtained from $\pi$ by removing $V_j$ from the ordering. To justify the equalities, we first note that terms of variables $V_i < V_j$ do not contain $V_j$ and can be brought outside the sum. 

To obtain the third equality, we multiply by $[\prod_{U \in \+ M} P(U|\+ E_U)]/[\prod_{U \in \+ M} P(U|\+ E_U)]$ and apply condition \eqref{eq:factorization} of Definition~\ref{def:simpsets} on the right-hand side as licensed by condition \eqref{eq:independence} of the definition. To obtain the fourth equality, we simply carry out the summation in terms of $V_j$. Conditions  \eqref{eq:factorization} and  \eqref{eq:independence} of Definition~\ref{def:simpsets} make it possible to refactorize the joint term into product terms so that the terms corresponding to variables $U \in \+ M$ remain unchanged and can be divided out once more. Thus we obtain the last equality, and an expression that no longer contains $V_j$ and has the same value as $A$.
\end{proof}

\section{Derivation of the Causal Effect in the Introductory Example}
\label{app:derivation}
We present the derivation of the causal effect of $X$ on $Y,Z_3,Z_2,Z_1$ in the graph $G$ of Figure~\ref{fig:grfG} using Algorithm~\ref{alg:identify}. We fix topological ordering of $G$ as $Z_2 < X < Z_1 < Z_3 < Y$. The original call $\Call{ID}{\{Y,Z_1,Z_2,Z_3\},\{X\},P(\+ V),G}$ fires line 4 and results in three new recursive calls. We have
\begin{equation} \label{eq:deriv-start}
P_{X}(Y,Z_3,Z_1,Z_2) = P_{Y,Z_3,X,Z_2}(Z_1)P_{Y,Z_1,X,Z_2}(Z_3)P_{Z_3,Z_1,X}(Y,Z_2),
\end{equation}
as the graph $G[\+ V \setminus \{X\}]$ has three C-components formed by the sets $\{Z_1\}$, $\{Z_3\}$ and $\{Y,Z_2\}$, respectively.

The first recursive call $\Call{ID}{\{Z_1\}, \{Y,Z_3,X,Z_2\}, P(\+ V), G}$ fires line 2 because $Z_3$ and $Y$ are not ancestors of $Z_1$. The next call $\Call{ID}{\{Z_1\}, \{X,Z_2\}, P(Z_1,X,Z_2), G[\{Z_1,X,Z_2\}]}$ fires line 6 because $C(G[\{Z_1\}])$ contains only one C-component and it is not part of a larger C-component in the graph of the current recursion stage. We have
\begin{equation} \label{eq:deriv-pt2}
P_{Y,Z_3,X,Z_2}(Z_1) = P_{X,Z_2}(Z_1) = P(Z_1|X,Z_2).
\end{equation}
To obtain $P_{Y,Z_1,X,Z_2}(Z_3)$ we call $\Call{ID}{\{Z_3\}, \{Y,Z_1,X,Z_2\}, P(\+ V), G}$ which also fires line 2 because $X$, $Z_1$ and $Y$ and not ancestors of $Z_3$. Calling $\Call{ID}{\{Z_3\}, \{Z_2\}, P(Z_3,Z_2), G[\{Z_3,Z_2\}]}$ fires line 6 $C(G[\{Z_3\}])$ contains only one C-component and it is not part of a larger C-component in the graph of the current recursion stage. We have
\begin{equation} \label{eq:deriv-pt1}
P_{Y,Z_1,X,Z_2}(Z_3) = P_{Z_2}(Z_3) = P(Z_3|Z_2).
\end{equation}
To obtain the last term we call $\Call{ID}{\{Y,Z_2\}, \{Z_3,Z_1,X\}, P(\+ V), G}$. The subgraph $G[\+ V \setminus \{Z_3,Z_1,X\}] = G[\{Y,Z_2\}]$ has only one C-component, but it is part of a larger C-component formed by the set $\+ S^\prime = \{Y,Z_3,X,Z_2\}$ in the current graph $G$. Line 7 is fired resulting in
\begin{equation} \label{eq:deriv-line7}
\Call{ID}{\{Y,Z_2\}, \{Z_3,X\}, P(Y|Z_3,Z_1,X,Z_2)P(Z_3|Z_2,X)P(X|Z_2)P(Z_2), G[\+ S^\prime]}.
\end{equation}
This call fires line 2 since $X$ is not an ancestor of $Y$ in the graph $G[\+ S^\prime]$. Letting $\+ T = \+ S^\prime \setminus \{X\} = \{Y,Z_3,Z_2\}$ the next call is
\begin{equation} \label{eq:deriv-line7cont}
\Call{ID}{\{Y,Z_2\}, \{Z_3\}, \sum_{X} P(Y|Z_3,Z_1,X,Z_2)P(Z_3|Z_2,X)P(X|Z_2)P(Z_2), G[\+ T]}.
\end{equation}
This time we trigger line 6 because $G[\+ T \setminus \{Z_3\}]$ has only one C-component and there is no larger C-component of $G[\+ T]$ that would contain it. We obtain
\begin{equation} \label{eq:deriv-line7prod}
  \begin{aligned}
P_{Z_3,Z_1,X}(Y,Z_2) &= P_{Z_3,Z_1,X}(Y,Z_2) \\
                     &= P_{Z_3,X}(Y,Z_2) \\
                     &= P_{Z_3}(Y,Z_2) \\
                     &= P^*(Y|Z_3,Z_2)P^*(Z_2),
  \end{aligned}
\end{equation}
where $P^*$ is the distribution of the current recursion stage, that is 
\[ P^*(Y,Z_3,Z_2) = \sum_{X} P(Y|Z_3,Z_1,X,Z_2)P(Z_3|Z_2,X)P(X|Z_2)P(Z_2). \]
In order to represent the conditional probability on the last line of \eqref{eq:deriv-line7prod}, we write
\begin{equation} \label{eq:deriv-line7quotient}
  \begin{aligned}
  P^*(Y|Z_3,Z_2)P^*(Z_2) &= \frac{P^*(Y,Z_3,Z_2)}{P^*(Z_3,Z_2)} P^*(Z_2) \\
    &= \frac{P^*(Y,Z_3,Z_2)}{\sum_{Y} P^*(Y,Z_3,Z_2)} \sum_{Y,Z_3} P^*(Y,Z_3,Z_2) \\
    &= \frac{\sum_{X}P(Y| Z_2,X,Z_3,Z_1)P(Z_3| Z_2,X)P(X| Z_2)P(Z_2)}{\sum_{X,Y}P(Y| Z_2,X,Z_3,Z_1)P(Z_3| Z_2,X)P(X| Z_2)P(Z_2)} \times \\
    &\quad\sum_{X,Z_3,Y}P(Y| Z_2,X,Z_3,Z_1)P(Z_3| Z_2,X)P(X|Z_2)P(Z_2).
  \end{aligned}
\end{equation}
Finally, we gather the results of our subproblems in \eqref{eq:deriv-pt2}, \eqref{eq:deriv-pt1} and \eqref{eq:deriv-line7quotient}, and insert them back into the equation in \eqref{eq:deriv-start} which yields
\begin{align*}
 P_{X}(Y,Z_3,Z_1,Z_2) &= P(Z_1| Z_2,X)P(Z_3| Z_2) \times \\
 &\frac{\sum_{X}P(Y| Z_2,X,Z_3,Z_1)P(Z_3| Z_2,X)P(X| Z_2)P(Z_2)}{\sum_{X,Y}P(Y| Z_2,X,Z_3,Z_1)P(Z_3| Z_2,X)P(X| Z_2)P(Z_2)} \times \\
 &\sum_{X,Z_3,Y} P(Y| Z_2,X,Z_3,Z_1)P(Z_3| Z_2,X)P(X| Z_2)P(Z_2)
\end{align*}
as the formula for the causal effect.

\section{Proof of Theorem~\ref{thm:complete}} 
\label{app:complete}

\begin{proof} (i) Suppose that $\Call{simplify}{A,G,\pi}$ has returned an expression with variable $V_j$ eliminated. Because the computation completed successfully, we have that each application of $\Call{join}{}$ and $\Call{insert}{}$ succeed. We can rewrite the value of $A$ as
\[   \prod_{V_i < V_j} P(V_i|\+ C_i) \sum_{V_j} \prod_{V_i \geq V_j} P(V_i|\+ C_i),  \]
where the terms $P(V_i|\+ C_i)$ such that $V_i < V_j$ can be brought outside the sum over $V_j$, because they cannot contain $V_j$. The functions $\Call{join}{}$ and $\Call{insert}{}$ use only standard rules of probability calculus, which can be seen on line~\ref{line:checkcond} of Algorithm~\ref{alg:join} and line~\ref{line:checkindep} of Algorithm~\ref{alg:insert}, and thus every new formation of a joint distribution $P(\+ J|\+ D)$ has been valid. Once again we rewrite the value of $A$ as
\[  \prod_{V_i < V_j} P(V_i|\+ C_i) \sum_{V_j} P(\+ J| \+ D),  \]
which means that condition \eqref{eq:factorization} of Definition~\ref{def:simpsets} is now satisfied, as we have obtained a joint term from the original product terms.Because $V_j \in \+ J$ we can carry out the summation which yields
\[  \prod_{V_i < V_j} P(V_i|\+ C_i) \cdot P(\+ J \setminus \{V_j\}| \+ D), \]
Because Algorithm~\ref{alg:simplify} succeeds, we know that  every insertion is canceled out by $\Call{factorize}{}$. To complete the procedure we obtain a new factorization without $V_j$ resulting in an atomic expression $A^\prime$ that no longer contains $V_j$. Condition \eqref{eq:independence} of Definition~\ref{def:simpsets} is satisfied by the definition of $\Call{insert}{}$, because the function always checks the conditional independence with the current summation variable on line~\ref{line:checkindep}.  Both conditions for simplification sets have been satisfied by construction.

(ii) Suppose that there exists a collection of simplification sets of $A$ with respect to $V_j$. For the sake of clarity, assume further that $V_n = V_j$. This assumption lets us only consider those terms that are relevant to the simplification of $V_j$, as we can always move conditionally independent terms outside the summation and consider only the expression remaining inside the sum. Let us first assume that $\+ M = \emptyset$. In this case condition \eqref{eq:factorization} simply reads
\[   \prod_{V_i \geq V_j} P(V_i | \+ C_i) = P(V_j,\ldots, V_1|\+ D), \]
and that the product terms are a factorization of the joint term. However, we want to show that they also provide a factorization that agrees with the topological ordering.
Because $A$ is $\pi-$consistent, for any two variables $V > W$ we have that $\+ C_W \subseteq V^\pi$ which enables us to consider the summations from $V_k$ up to $V_1$ for $k = 1,\ldots,j-1$, which results in
\[  \sum_{V_k,\ldots,V_1}  \prod_{V_i \geq V_j}  P(V_i | \+ C_i) = \sum_{V_k,\ldots,V_1} P(V_j,\ldots, V_1|\+ D) = P(V_j,\ldots, V_{k+1}|\+ D). \]
We obtain for $k = j-1,\ldots,1$
\begin{gather}  \label{eq:manyeqs}
\begin{aligned}
P(V_j|\+ C_j) &= P(V_j|\+ D) \\
P(V_j|\+ C_j)P(V_{j-1}|\+ C_{j-1}) &= P(V_j, V_{j-1}|\+ D) \\
                        &\;\;\vdots \\
P(V_j|\+ C_j)\cdots P(V_2|\+ C_2) &= P(V_j,\ldots, V_2|\+ D) \\
P(V_j|\+ C_j)\cdots P(V_2|\+C_2)P(V_1|\+ C_1) &= P(V_j,\ldots, V_1|\+ D).
\end{aligned}
\end{gather}
From the last and second to last equation we can obtain 
\[  P(V_j,\ldots, V_2|\+ D)P(V_1|\+ C_1) =  P(V_j,\ldots, V_1|\+ D), \]
and by dividing with the first term from the left hand side we obtain
\[  P(V_1|\+ C_1)  = P(V_1|V_j,\ldots,V_2,\+ D). \]
In fact, we can do this for any two subsequent equations in $\eqref{eq:manyeqs}$ to obtain
\[  P(V_i|\+ C_i)  = P(V_i|V_j,\ldots,V_{i+1},\+ D), \quad i = 1,\ldots,j-1 \]

Algorithm~\ref{alg:simplify} operates by starting from $V_1$, so we still have to show it succeeds in constructing the joint term. Using the previous results we can rewrite the original equation as
\[   \prod_{V_i \geq V_j} P(V_i | \+ C_i) = \prod_{V_i \geq V_j} P(V_i | \+ C_i^*), \]
where $\+ C_i^* = \+ D \cup \{V_j, \ldots, V_{i+1}\}$ for $i < j$ and $\+ C_j^* = \+ D$. From this we obtain
\begin{gather}  \label{eq:manyeqs2}
  \begin{aligned}
P(V_1|\+ C_1) &= P(V_1|\+ C_1^*) \\
P(V_1|\+ C_1^*)P(V_2|\+ C_2) &= P(V_1, V_2|\+ C_2^*) \\
                        &\;\;\vdots \\
P(V_1,\ldots, V_{j-1}|\+ C_{j-1}^*)P(V_j|\+ C_j) &= P(V_j,\ldots, V_1|\+ C_j^*).
  \end{aligned}
\end{gather}
The function \Call{join}{} will succeed every time since the for-loop starting on line~\ref{line:forloop1} of Algorithm~\ref{alg:join} will discover the conditional independence properties allowing the previous equalities in \eqref{eq:manyeqs2} to take place. Thus Algorithm~\ref{alg:simplify} will return an atomic expression with the variable $V_j$ eliminated from the summation set. 

Assume now that $\+ M \neq \emptyset$ and let $\+ V = V[A]$ and. In this case condition \eqref{eq:factorization} allows us to write
\[   \prod_{U \in \+ M} P(U|\+ E_U) \prod_{V_i \geq V_j} P(V_i | \+ C_i) = P(\+ V, \+ M|\+ D), \]
and furthermore, we have that these product terms are a factorization of the joint term. First, we aim to reduce the number of variables in $\+ M$ to be considered. This is done because Algorithm~\ref{alg:simplify} always starts and finishes the construction of the joint term with a variable in $\+ V$. We categorize each $U \in \+ M$ into three disjoint sets. We define 
\[ \+ M^{-} := \{U \in \+ M \mid U \not\in \bigcup_{k = 1}^j \+ C_k \}\;, \+ M^{+} := \{U \in \+ M \mid U \in \bigcap_{k = 1}^j \+ C_k\}  \textrm{ and } \] 
\[  \+ M^* :=  \+ M \setminus (\+ M^{-} \cup \+ M^{+}). \]
 First, we show that we can ignore variables in $\+ M^{-}$ by obtaining a new factorization without them. It follows from the definition of $\+ M^{-}$ and \eqref{eq:factorization} that we can compute the marginalization as follows
\begin{align*}
 P(\+ V, \+ M \setminus \+ M^{-}|\+ D) &= \sum_{U \in \+ M^{-}}  P(\+ V, \+ M|\+ D)  \\
                                       &= \sum_{U \in \+ M^{-}} \prod_{U \in \+ M} P(U | \+ E_U) \prod_{V_i \geq V_j} P(V_i | \+ C_i)  \\
                                       &= \prod_{V_i \geq V_j} P(V_i | \+ C_i) \sum_{U \in \+ M^{-}} \prod_{U \in \+ M} P(U | \+ E_U)  \\
                                       &= \prod_{U \in \+ M \setminus \+ M^{-}} P(U | \+ E_U) \prod_{V_i \geq V_j} P(V_i | \+ C_i).
\end{align*}
We have a new factorization without any variables in $\+ M^{-}$. Similarly, we can eliminate the variables in $\+ M^{+}$ from our factorization. It follows from the definition of $\+ M^{+}$ that for all $U \in \+ M^{+}$ we have that $\+ E_U \subseteq \+ D$. From this we obtain
\[  \prod_{U \in \+ M^{+}} P(U|\+ E_U) = P(\+ M^{+}|\+ D). \]
We can now write
\begin{align*}
P(\+ V, \+ M^* | \+ D, \+ M^{+}) &= \frac{P(\+ V, \+ M \setminus \+ M^{-})}{P(\+ M^{+}|\+ D)} \\
                                                       &= \frac{\prod_{U \in \+ M \setminus \+ M^{-}} P(U | \+ E_U) \prod_{V_i \geq V_j} P(V_i | \+ C_i)}{\prod_{U \in \+ M^{+}} P(U|\+ E_U)} \\
&= \prod_{U \in \+ M^*} P(U | \+ E_U) \prod_{V_i \geq V_j} P(V_i | \+ C_i).
\end{align*}
Thus it suffices to consider the factorization given by
\begin{equation} \label{eq:factorization2}
\prod_{U \in \+ M^*} P(U | \+ E_U) \prod_{V_i \geq V_j} P(V_i | \+ C_i) = P(\+ V, \+ M^*|\+ D^*),
\end{equation}
where $\+ D^* = \+ D \cup \+ M^{+}$. 

Next, we will order the variables in $\+ M^*$. For each $U \in \+ M^*$ we find the largest index $u \in \{1,\ldots,j-1\}$ such that $U \in \+ C_u$. This choice is well defined, since by definition at least one such index exists. Furthermore, as the product terms in \eqref{eq:factorization2} are a factorization of the joint term, the conditioning sets are increasing and we have that $U \not\in \+ C_i $ for all $i \geq u+1$. In the case that multiple variables $U_i \in \+ M^*$ for some set of indices $i \in \+ I$ share the same index $u$, we may redefine $M^*$ such that $U_i, i \in \+ I$ are replaced by a single variable $U_I$ such that $\prod_{i \in I} P(U_i|\+ E_{U_i}) = P(U_I|\+ E_{U_I})$, where $\+ E_{U_I} = \cap_{i \in \+ I} \+ E_{U_i}$. Thus we can assume that for any two variables $U_1, U_2 \in \+ M^*$ we have that $u_1 \neq u_2$. We can now order the variables in $\+ M^*$ by their respective indices $u$ such that $U_1 > U_2 > \ldots > U_m$ and $u_1 < u_2 < \ldots < u_m$. 

Nest we will extend the ordering to include all of the variables in the set $\+ V$. We let $\+ Q := \+ V \cup \+ M^*$ and find an ordering of this set such that it agrees with induced ordering $\omega$ of the variables in $\+ V$ and with the ordering of the indices $u_1,\ldots,u_m$. A new factorization given by this ordering can be defined as follows:
$$ Q_{k} = \begin{cases}
V_{k-m} & k > u_m, \\
V_{k-l} & u_{l} < k < u_{l+1}, \\
V_{k} & k < u_1, \\
U_l & k = u_l.
\end{cases} \quad 
 \+ D_k = \begin{cases}
\+ C_{k-m} & k > u_m, \\
\+ C_{k-l} & u_{l} < k < u_{l+1}, \\
\+ C_{k} & k < u_1, \\
\+ E_{U_l} & k = u_l.
\end{cases}
$$
We can now rewrite the factorization of \eqref{eq:factorization2} as
\begin{equation} \label{eq:factorization3}
\prod_{k = 1}^{n+m} P(Q_k | \+ D_k) = P(\+ Q|\+ D^*),
\end{equation}
We can now apply the same procedures as in the case of $\+ M = \emptyset$ with the exception that $\Call{insert}{}$ succeeds where $\Call{join}{}$ fails with terms containing $Q_k$ and $Q_{k+1}$ when $k = l-1$ for all $l = 1,\ldots,m$. The success of $\Call{insert}{}$ is guaranteed by condition \eqref{eq:independence}, as the function will find this conditional independence on line~\ref{line:checkcond} of Algorithm~\ref{alg:insert}. Also, $\Call{factorize}{}$ will remove all additional terms that were introduced in the process, which is made possible by condition \eqref{eq:independence} and the definition of the factorization of $P(\+ Q|\+ D^*)$. After the summation over $V_j$ is carried out, the conditional independence between $V_j$ and the variables $U \in \+ M^*$ ensures that their respective terms are equal to the original factorization before the summation was carried out when the new factorization is constructed so that it agrees with the ordering of the set $\+ Q$. Thus an atomic expression is returned with the variable $V_j$ eliminated with the same value as the original atomic expression.
\end{proof}

\bibliography{simplification}

\end{document}